\documentclass{article}
\usepackage{iclr2027_conference,times}

\usepackage{hyperref, url}
\usepackage[table]{xcolor}
\definecolor{avgshade}{gray}{0.92}
\definecolor{meanbg}{HTML}{D6F0FE}
\usepackage{graphicx}
\usepackage{booktabs}
\usepackage{amsmath}
\usepackage{amssymb}
\usepackage{mathtools}
\usepackage{float}

\usepackage{xspace}

\title{BrainCodes: A Lightweight EEG Foundation \\ Model over Discrete Neural Codes}

\author{Kleanthis Avramidis\thanks{Work done during an internship at Microsoft Research.} \\
Signal Analysis and Interpretation Lab \\
University of Southern California \\
Los Angeles, CA 90007, USA \\
\texttt{avramidi@usc.edu} \\
\And
Ivan Tashev \& Dimitra Emmanouilidou \\
Microsoft Research \\
Redmond, WA, USA \\
\texttt{ivantash@microsoft.com}\\
\texttt{diemmano@microsoft.com} 
}

\iclrfinalcopy 

\begin{document}

\maketitle
\lhead{Preprint. Under review.}

\begin{abstract}
Recent electroencephalography (EEG) foundation models increasingly emphasize model scale, potentially limiting their deployment on resource-constrained devices and in everyday environments. We argue that transferable EEG features can instead be learned with a compact architecture and introduce \textbf{BrainCodes}, a lightweight foundation model pretrained through self-supervised learning on more than 800 hours of recordings from over 1,000 participants. Our architecture combines a quantization-based neural tokenizer with a spatiotemporal encoder that supports arbitrary electrode configurations. The pretraining objective explicitly encourages representations that capture spatial connectivity through channel positional encoding, and sensitivity to narrowband neural oscillations through an auxiliary training objective. Our model matches or surpasses the state of the art across diverse neurophysiological monitoring tasks, including sleep staging, attention assessment, and cognitive load estimation. BrainCodes transfers effectively to low-resource wearable devices while using up to 97\% fewer trainable parameters than comparable architectures. Beyond predictive accuracy, our evaluation framework characterizes the geometry and invariances of learned EEG representations across measurable spatial, temporal, and spectral signal properties.
\end{abstract}

\section{Introduction}

Recent self-supervised models for electroencephalography (EEG) are typically trained on patch reconstruction, either in the signal waveform or in a spectral representation~\citep{labram,cbramod,biot,reve}. This objective is inherited from language and computer vision, where a masked token can be reliably estimated from its local context. EEG however presents a different setting, in which a masked interval is only weakly constrained by its neighborhood, while neighboring electrodes often contain redundant mixtures of the same underlying sources due to volume conduction. Consequently, components that carry task-relevant properties may occupy small fraction of the total variance, so that a model need not preserve them to reduce reconstruction error. Whether common EEG pretext tasks produce representations grounded in those properties is an empirical question that downstream benchmarks alone can only partially answer.

Consider reconstruction under a squared-error objective in the frequency domain. EEG typically exhibits an aperiodic power spectrum $S(f) \propto f^{-\alpha}$~\citep{he2014,fooof}, and the contribution of a frequency band to the reconstruction loss depends on its residual power. High-power broadband components can therefore dominate the objective, whereas narrowband, oscillatory components may contribute relatively little to the optimization process. A pretrained model may consequently achieve low reconstruction error by accurately fitting only the aperiodic background, without faithfully preserving rhythmic structure. Spatial reconstruction presents an analogous ambiguity. Scalp potentials are linear lead-field mixtures of underlying source currents, and their spatial autocorrelation is shaped not only by source dynamics but also by head geometry and electrode density~\citep{nunez}. Because neighboring electrodes often contain redundant mixtures, a masked channel can be approximated accurately from its neighbors; even spherical spline interpolation can recover missing signals with high fidelity~\citep{perrin1989}. Masked-channel reconstruction may therefore be solved partly through spatial interpolation without requiring the representation to encode physiologically informative spatial dynamics.

Although these ambiguities are well established in neurophysiology, standard downstream benchmarks do not directly determine which properties of EEG signals are retained by a foundation model. We therefore introduce a compact model, trained to predict a finite set of discrete latent codes produced by a vector-quantized tokenizer. The quantizer allocates capacity according to a rate–distortion trade-off, concentrating representational capacity on distinctions between samples. Because this prediction objective is categorical rather than a squared error on the waveform level, its weighting is not directly governed by absolute signal power. To promote sensitivity to physiologically meaningful oscillatory and spatial structure, we supplement masked-token prediction with an auxiliary objective based on localized narrowband perturbations, and we explicitly encode relative electrode distances as positional encoding. The central novelty is therefore an objective designed to preserve oscillatory and spatial structure without relying on scale. Our contributions are as follows:

\begin{itemize}
\item \textbf{A compact foundation model on top of discrete codes.} BrainCodes combines a vector-quantized tokenizer and a series of factorized transformer layers trained on masked code prediction under an oscillation-injection auxiliary. With 1.9M trainable parameters in its backbone, our model achieves the highest mean performance among the compared models across 14 downstream benchmarks in both linear probing and full fine-tuning.
 
\item \textbf{Transfer to sparse, consumer-grade montages.} To our knowledge, we provide the first systematic evaluation of EEG foundation models on consumer-grade hardware, using three datasets spanning in-ear EEG and the Muse headband. BrainCodes substantially outperforms the published baselines in these settings. Through controlled sparse-montage simulations, we further demonstrate that BrainCodes retains, on average, 91\% of its full-montage performance when using only a sparse set of 4 electrodes.
 
\item \textbf{Evidence of retained spatial and spectral structure.} We introduce controlled probes that measure representational sensitivity to electrode geometry and spectral perturbations. Relative to the compared models, BrainCodes exhibits stronger geometry-consistent spatial locality and greater sensitivity to oscillatory components than to the aperiodic slope of the EEG spectrum. These results demonstrate that its representations retain measurable spatial and spectral properties of EEG beyond those captured by downstream performance alone.

\end{itemize}

\section{Methods}
\label{sec:methods}

We introduce \textbf{BrainCodes}, a self-supervised EEG model designed to preserve measurable spatial and spectral properties of scalp-recorded signals in a reduced parameter budget. A frozen RVQ tokenizer first converts each EEG channel into continuous embeddings associated with discrete code indices~\citep{biocodec}, and then a hierarchical factorized transformer is trained over those tokens under a masked-reconstruction primary objective and an oscillation-prediction auxiliary. A transformer U-Net backbone with factorized temporal and channel attention processes the RVQ embeddings and predicts discrete code indices at masked positions. Auxiliary heads applied to its time-compressed bottleneck predict the frequency, duration, and spatial extent of synthetic narrowband perturbations. The resulting bottleneck features are used for all downstream tasks.

In the following, let $X \in \mathbb{R}^{C \times T_s}$ denote an EEG window with $C$ channels and $T_s$ samples at given sampling rate, and let $\mathbf{p}_c \in \mathbb{R}^3$ be the standard montage coordinate of electrode $c$. 

\subsection{Discrete Channel Tokenization}

Each EEG channel is encoded independently by a pretrained tokenizer $\mathcal{E}$, held frozen throughout training. The tokenizer, realized as the publicly available BioCodec model~\citep{biocodec}, quantizes a single-channel waveform with an RVQ of $K=6$ stages over a shared vocabulary of size $V=256$, each stage selecting an entry of a low-dimensional codebook of $d_c=16$,
\begin{equation}
\begin{gathered}
  Q_c = \big(q_{c,t,k}\big) \in \{1,\dots,V\}^{K \times T},
  \qquad q_{c,t,k}\in\{1,\dots,V\},\\
  \mathbf{z}_{c,t} \;=\; \sum_{k=1}^{K} \mathbf{W}_k\, \mathbf{e}_k\!\left(q_{c,t,k}\right)
  \;\in\; \mathbb{R}^{D},
  \qquad D = 128 ,
\end{gathered}
\label{eq:vq}
\end{equation}
Here, $q_{c,t,k}$ is the codebook index for channel $c$, frame  $t$, and RVQ stage $k$. $\mathbf{z}_{c,t}$ denotes code embeddings, $\mathbf{e}_k(\cdot)$ is the $k$-th codebook and $\mathbf{W}_k$ its output projection. The interface to the downstream model is the aggregate $D=128$ embedding obtained by summing the dequantized residuals,
\begin{equation}
  \mathcal{E}: \mathbb{R}^{T_s} \rightarrow \mathbb{R}^{D \times T},
  \qquad
  Z_c = \mathcal{E}(X_c) = \big(\mathbf{z}_{c,1},\dots,\mathbf{z}_{c,T}\big),
\end{equation}
over $T$ code frames. The model thus operates on continuous representations while the discrete RVQ codes define the bottleneck. Because tokenization is performed independently for each channel and channel attention operates over a variable axis, the architecture can accommodate different electrode sets, provided that a spatial coordinate is available for each electrode.

\subsection{Spatiotemporal Positional Encoding}

\begin{figure}[ht]
    \centering
    \includegraphics[width=\linewidth]{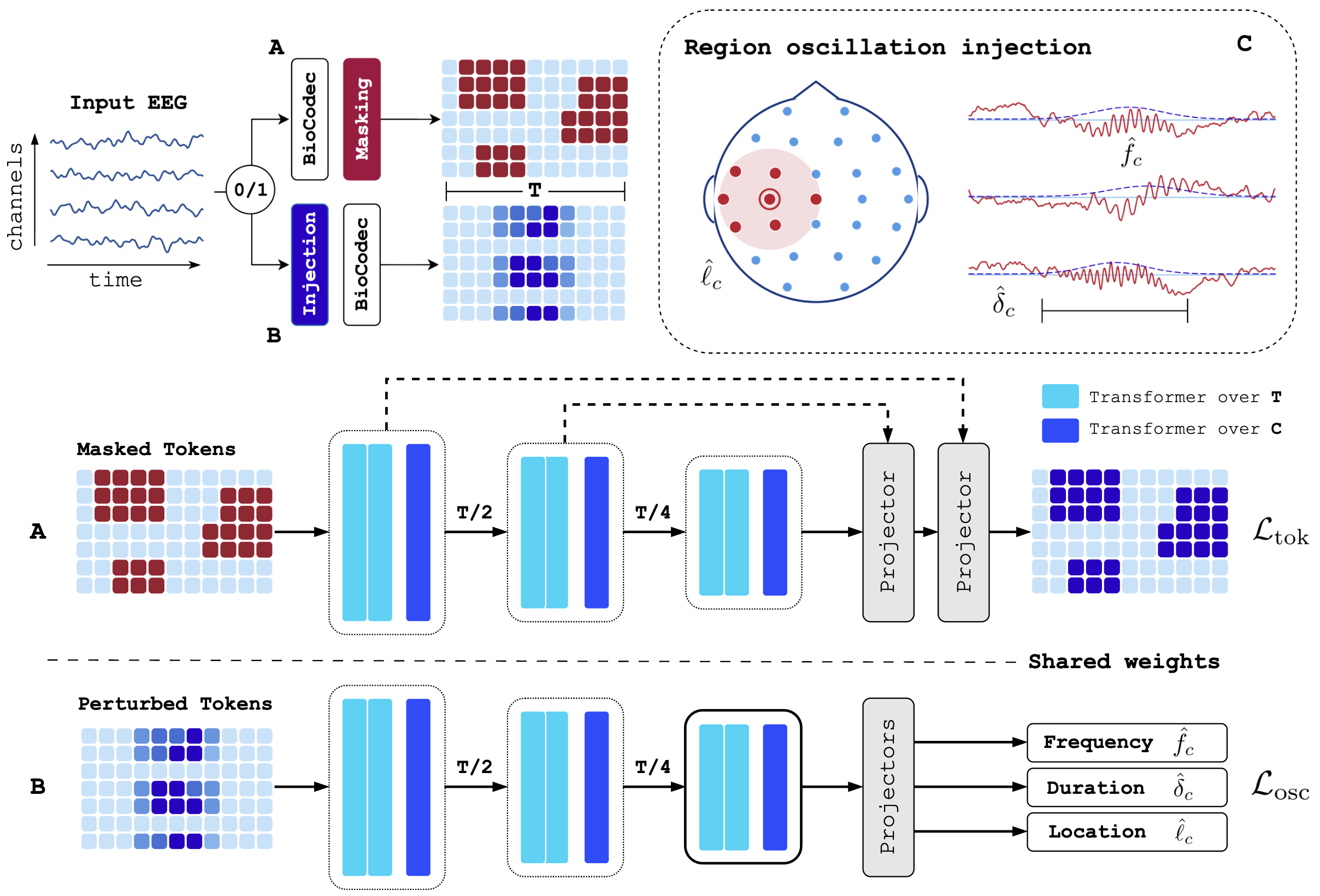}
    \vspace{-5mm}
    \caption{\textbf{The pretraining objectives of BrainCodes}. \textbf{A)} Contiguous blocks of codes are masked and the model predicts the missing indices, evaluated through mean cross-entropy over the masked positions ($\mathcal{L}_{\mathrm{tok}}$). \textbf{B)} Narrowband bursts are injected into regions of electrodes and the bottleneck is read by linear heads for its frequency, duration and location ($\mathcal{L}_{\mathrm{osc}}$).}
    \label{fig:training}
    \vspace{-2mm}
\end{figure}

The token at position $(c,t)$ is the aggregate code embedding $\mathbf{z}_{c,t}\in\mathbb{R}^{D}$ of Eq.~\ref{eq:vq}. The backbone factorizes self-attention into a temporal axis, applied over $t$ for each channel, and a channel axis, applied over $c$ at each time step. The two axes are encoded separately: time is an ordered one-dimensional index, whereas channels are embedded in continuous three-dimensional space. Temporal position is injected inside the temporal-attention modules via standard RoPE~\citep{su2024roformer}.

Each channel label is resolved against the MNE \texttt{standard\_1005} montage to a physical coordinate $\mathbf{p}_c\in\mathbb{R}^3$. Coordinates are centered on the mean of the selected electrodes and divided by the largest centered norm, giving $\tilde{\mathbf{p}}_c$ in the unit ball. Absolute placement is therefore removed, while relative distances are preserved. We encode $\tilde{\mathbf{p}}_c$ with a mixed-axis rotary formulation following~\citet{heo2024rotary}, extended from 2D grids to continuous three-dimensional coordinates. Let $H$ be the number of heads, $d_h=D/H$ the head dimension, and $R=d_h/2$ the number of rotary pairs. Each head $h$ and pair $r$ carries a learnable frequency vector $\boldsymbol{\omega}_{h,r}\in\mathbb{R}^{3}$, so the rotation angle is
\begin{equation}
\theta_{h,c,r}=\boldsymbol{\omega}_{h,r}^{\top}\tilde{\mathbf{p}}_c .
\end{equation}
Compared to the baseline axial variant, the mixed formulation lets every subspace select its own orientation during pretraining, thereby fitting to anatomically meaningful axes of scalp activations.

The frequency vectors are initialized with log-spaced magnitudes $\lVert\boldsymbol{\omega}_{h,r}\rVert=\text{base}^{-r/R}$ and random unit directions, both of which are optimized during pretraining. Angles are duplicated across the two halves of the head dimension and applied to the queries and keys $QK$ as
\begin{equation}
R_{h,c}(\mathbf{v})=\mathbf{v}\odot\cos\boldsymbol{\Theta}_{h,c} +\mathrm{rot}_{1/2}(\mathbf{v})\odot\sin\boldsymbol{\Theta}_{h,c}\,,
\end{equation}
with $\mathrm{rot}_{1/2}([\mathbf{v}^{(1)},\mathbf{v}^{(2)}])=[-\mathbf{v}^{(2)},\mathbf{v}^{(1)}]$. Since $R(\theta_i)^{\top}R(\theta_j)=R(\theta_j-\theta_i)$ and $\theta$ is linear in
$\tilde{\mathbf{p}}_c$, the positional part of the attention logit reduces to
\begin{equation}
\Delta\theta_{h,i,j,r}=
\boldsymbol{\omega}_{h,r}^{\top}(\tilde{\mathbf{p}}_j-\tilde{\mathbf{p}}_i).
\end{equation}
Effectively, channel attention is informed exclusively by relative electrode displacement rather than by absolute positions, so that any offset shared by all electrodes would cancel in the model's latent representation. During pretraining we also perturb coordinates with Gaussian jitter whose standard deviation is capped at half the mean nearest-neighbor spacing of the montage.

The reason to encode displacement rather than absolute coordinates is that displacement is what carries information at the sensor level: volume conduction and inter-electrode correlation depend on how far apart and in which direction two electrodes sit, not on where either lies in an external frame. Absolute position is instead deferred to fine-tuning, where montages are fixed.

\subsection{Factorized U-Net Backbone}
\label{sec:model}

The U-Net backbone $f_\theta$ is a stack of three transformer blocks, each comprising three encoder layers. Within a block, the first two layers apply \emph{time-mixing} attention over $t$ within a channel, while the third applies \emph{channel-mixing} attention over $c$ at fixed $t$. Factorizing attention has been validated in prior work~\citep{sttransformer,cbramod,csbrain} and avoids the expensive operation of attending jointly over all $C \times T$ tokens.

Following the hourglass design of U-Nets, the encoder pools along time by factors $\{p_i\}$, contracting to a bottleneck of length $T_b = T / \prod_i p_i$ (here $\prod_i p_i = 4$), and a symmetric decoder upsamples back to full resolution, adding skip connections $S_j$ under dropout. Skip dropout is used to discourage the decoder from bypassing the bottleneck, forcing the compressed representation to retain
reconstructive information. The model architecture is shown in Figure~\ref{fig:training}. Let the normalized bottleneck features be $B\in\mathbb{R}^{C\times T_b\times D}$, with $B_{c,t}\in\mathbb{R}^{D}$. We derive two downstream representations from $B$:
\begin{enumerate}
    \item its channel-wise time-averaged $\,\mathbf{b}_c = \frac{1}{T_b}\sum_{t=1}^{T_b} B_{c,t} \,\in\, \mathbb{R}^{D},$
    \item its channel-wise flattened $\,\bar{\mathbf{b}}_c = \big[\,B_{c,1};\dots;B_{c,T_b}\,\big] \;\in\; \mathbb{R}^{T_b D}.$
\end{enumerate}
The time-averaged representation removes the explicit temporal axis, whereas the flattened one retains the ordering of the bottleneck time steps at the cost of a $T_b$-fold increase in feature dimension. Time averaging thus removes explicit access to event timing (\emph{when} an event occurs in the EEG interval), even though timing information may remain encoded in the averaged activations. We implement both in accordance with prior work~\citep{reve} showing that different benchmark datasets behave differently under each of those regimes.

\subsection{Pretraining Objectives}

\textbf{Masked code prediction.} A block mask selects a set $\mathcal{M}$ of codes $(c,t)$. The masked positions form continuous temporal intervals that cover 20--80\% of the selected channel's code frames. For each RVQ state, a linear head $g_k$ predicts the corresponding discrete code index $q_{c,t,k}$ from the full-resolution backbone state $\mathbf{h}_{c,t}$.  The loss is the mean cross-entropy (CE) over the masked positions:
\begin{equation}
  \mathcal{L}_{\text{tok}}
  = \frac{1}{|\mathcal{M}|}\sum_{(c,t)\in\mathcal{M}}
    \frac{1}{K}\sum_{k=1}^{K}
    \mathrm{CE}\,\big(g_k(\mathbf{h}_{c,t}),\, q_{c,t,k}\big).
\end{equation}
\textbf{Region oscillation injection.} To encourage the bottleneck to retain measurable spectral and spatial properties, we injected a synthetic narrowband burst into a spatially contiguous region of electrodes. A seed electrode $s$ was drawn uniformly and the region $\mathcal{R}$ was defined as its $\lceil \rho\, C\rceil$ nearest neighbors. For channel $c\in\mathcal{R}$ we added a Hann-windowed burst of center $f_c$, duration $\delta_c$, and amplitude $a_c$,
\begin{equation}
  \tilde{X}_c(t) = X_c(t) + a_c\,
  \underbrace{w\,\big(t;\tau_c,\delta_c\big)}_{\text{Hann envelope}}\,
  \underbrace{\eta_c\,\big(t; f_c,\Delta\big)}_{\text{band-limited carrier}},
\end{equation}
and re-tokenize only the affected channels. Using the bottleneck feature $\mathbf{b}_c$, auxiliary linear heads regress the min--max normalized frequency and duration of the injected burst, while an auxiliary location head $\psi_\ell$ predicts for each channel whether it was injected or not. The first two were trained under squared error and the last with BCE, giving $\mathcal{L}_{f}$, $\mathcal{L}_{\delta}$, and $\mathcal{L}_{\ell}$ respectively to the final objective:
\begin{equation}
  \mathcal{L}
  = \lambda_{\text{tok}}\,\mathcal{L}_{\text{tok}}
  + \lambda_{\text{aux}}\left(\mathcal{L}_{f} + \mathcal{L}_{\delta} + \mathcal{L}_{\ell}\right).
\end{equation}
Applying $\mathcal{L}_{f},\mathcal{L}_{\delta},\mathcal{L}_{\ell}$, at the bottleneck layer provides direct supervision for retaining the frequency, duration, and spatial extent of the injected perturbations in the downstream representation.

\section{Experimental Setup}
\label{sec:setup}

\begin{table}[t]
\caption{EEG datasets used for downstream inference. ERP stands for event-related potentials.}
\label{tab:datasets}
\begin{center}
\small
\setlength{\tabcolsep}{4pt}
\begin{tabular}{p{3cm}ccccrr}
\toprule
\textbf{Dataset} & \textbf{Category} & \textbf{Channels} & \textbf{Length (s)} & \textbf{Classes} & \textbf{Samples} & \textbf{Participants} \\
\midrule
\textit{Main Bench}\vspace{1mm} \\
TUAB           & Clinical  & 21 & 10 & 2 & 409,455 & 2,329 \\
TUEV           & Events & 16 & 5  & 6 & 106,394 & 370   \\
Siena          & Seizures  & 29 & 10 & 2 & 50,749  & 14    \\
PhysioNet-MI   & Motor     & 64 & 4  & 4 & 9,838   & 109   \\
BCI IV-2a      & Motor     & 22 & 4  & 4 & 5,184   & 9     \\
EEGMAT         & Stress    & 20 & 5  & 2 & 1,707   & 36    \\
ISRUC          & Sleep     & 6  & 30 & 5 & 89,283  & 100   \\
\midrule
\textsc{NeuroRVQ} \textit{Bench}\vspace{1mm} \\
High-Gamma     & Motor     & 78 & 4  & 4 & 13,484  & 14    \\
OpenBMI-ERP    & ERP       & 62 & 1  & 2 & 149,040 & 54    \\
Pavlov-Memory  & Memory    & 63 & 4  & 2 & 3,505   & 65    \\
PhysioNet-Eyes & Resting   & 64 & 4  & 2 & 3,270   & 109   \\
\midrule
\textit{Low Resource}\vspace{1mm} \\
EESM23         & Sleep     & 4  & 30 & 6 & 14,510  & 10    \\
Muse-Motor     & Motor     & 4  & 4  & 3 & 6,118   & 19    \\
Muse-Mental    & Cognitive & 4  & 4  & 3 & 296     & 4     \\
\bottomrule
\end{tabular}
\end{center}
\vspace{-5mm}
\end{table}

\subsection{Pretraining Corpus}

BrainCodes is pretrained on the Healthy Brain Network EEG dataset (HBN-EEG)~\citep{alexander2017open}, a developmental corpus of high-density recordings from children and adolescents acquired across multiple sites under a common protocol. We pretrain on a subset of HBN-EEG, comprising releases R3, R4, R7 and R8, which include about 820 hours from 1,131 participants. We chose this corpus because of its multiple axes of variability (age-wise, task-wise, and site-wise), and because of its dense 128-channel montage that allows training an extensive range of channel subsets.

Each recording was preprocessed in accordance with the requirements of the frozen tokenizer. We band-pass filtered the waveforms between 0.5 and 75\,Hz with a 60\,Hz notch, interpolated flat channels from their closest neighbors, resampled to $250$\,Hz, and robust-z-scored each channel independently. The only departure from the configuration of~\citet{biocodec} was the low-pass cutoff, which we moved from $100$ to $75$\,Hz as it improved convergence without regressions in reconstruction quality. Recordings were segmented into 30\,s windows and tokenized once with BioCodec. The tokenizer emits a 128D embedding from the RVQ output at a code-frame rate of $21$\,Hz, so a $30$\,s window yields $T=625$ frames. Windows of randomized length (2 to 30\,s) and channel set were drawn during training, for which we split a 20\% participant-disjoint validation set.

\subsection{Downstream datasets}

We evaluate our model on a total of 14 downstream tasks spanning three benchmark regimes (Table~\ref{tab:datasets}). The first is a standard suite used across the EEG foundation-model literature: abnormality detection and event classification on the Temple University corpora, TUAB and TUEV~\citep{obeid2016temple}; sleep staging on ISRUC~\citep{khalighi2016isruc}; seizure detection on Siena Scalp Database~\citep{detti2020eeg}; motor imagery on PhysioNet-MI~\citep{schalk2004bci2000} and BCI Competition IV-2a~\citep{brunner2008bci}; and cognitive load on the mental-arithmetic dataset~\citep{zyma2019electroencephalograms}. The second is the \textsc{NeuroRVQ} benchmark suite~\citep{neurorvq}, which adds High-Gamma motor~\citep{schirrmeister2017deep}, OpenBMI-ERP~\citep{lee2019eeg}, working memory~\citep{pavlov2022pupillometry} and eyes-open/closed resting~\citep{schalk2004bci2000}, presented in Appendix~\ref{app:neurorvq}. The third targets consumer-grade hardware: in-ear-EEG sleep staging on EESM23~\citep{bjarke2025ear} and two tasks recorded with the four-electrode Muse headband~\citep{elkaffas2024musemotor,bird2018study}, which together let us test transfer to montages an order of magnitude sparser than the pretraining layout.
 
Each dataset is processed by the same pipeline as the pretraining corpus, and electrode coordinates are resolved from the same standard montages, allowing one model to be fine-tuned across layouts of arbitrary number of electrodes without architectural changes. Splits are grouped so that no participant contributes to more than one partition. A representative selection of the aforementioned datasets is further used to quantify linear-probing robustness and the interpretability study.
 
\subsection{Implementation details}

\textbf{Baselines.} We compare against five models, namely BIOT~\citep{biot}, LaBraM~\citep{labram}, CBraMod~\citep{cbramod}, CSBrain~\citep{csbrain} and REVE~\citep{reve}, spanning $3.2$M to $69$M parameters. We additionally compare against EEGPT~\citep{eegpt}, NeuroGPT~\citep{neurogpt}, and \textsc{NeuroRVQ}~\citep{neurorvq} on the second benchmark (Appendix~\ref{app:neurorvq}). On the standard suite and the \textsc{NeuroRVQ} benchmark we quote published numbers where a directly comparable protocol exists. The wearable benchmark has no extensive published baselines, hence we adopt those from~\citet{tfmtokenizer} and additionally rerun the same models with the recipe used for BrainCodes.

\textbf{Hyperparameters.} All downstream runs optimized on AdamW with a cosine-decay schedule and linear warmup, weight decay 0.01, label smoothing 0.1, and square-root inverse-frequency class weights. Peak learning rate was chosen between $10^{-4}$ and $5\times10^{-4}$ across datasets, and training lasted 20 epochs for the majority of tasks (10 on TUAB, TUEV). Model selection was by balanced accuracy on the validation set. Every reported number is the mean over 3 seeds; cross-validation entries are additionally averaged over 4 folds in Section~\ref{sec:wearable} and 5 folds in Appendix~\ref{app:neurorvq}.

\textbf{Evaluation.} Both bottleneck features $\mathbf{b}_c$ and $\bar{\mathbf{b}}_c$ are assessed under two fine-tuning protocols. Under \emph{full fine-tuning}, the U-Net backbone (excluding the tokenizer) is optimized end to end along with a linear head that reads $\{\mathbf{b}_c\}_{c=1}^{C}$ or $\{\bar{\mathbf{b}}_c\}_{c=1}^{C}$. This protocol establishes the benefit from the pretrained initialization. Under \emph{linear probing}, the backbone is held frozen and a single linear head $\varphi$ is fitted to the same bottleneck features. Any performance recovered in this regime is thus attributable to the pretrained encoder, which makes it the most representative test of representation quality. For a $\mathcal{C}$-class downstream task we match the convention of the domain literature and report balanced accuracy (BAC) as the primary metric, together with AUROC and AUPRC when $\mathcal{C} = 2$, or Weighted F1 and Cohen's $\kappa$ when $\mathcal{C} > 2$. To render results comparable across tasks with differing numbers of classes, we report a fair \textit{average} model performance through chance-normalized BAC, i.e.,
\begin{equation}
    \mathrm{nBAC} = \left(\mathrm{BAC} - 1/\mathcal{C}\right) / \left(1 - 1/\mathcal{C}\right),
\end{equation}
which maps chance-level performance to $0$ and perfect classification to $1$.

\section{Results}
\label{sec:results}

\subsection{Full Fine-tuning Performance}

\begin{table}[t]
\caption{Balanced accuracy on the main-bench datasets under full fine-tuning, mean $\pm$ s.d. over seeds. \textbf{Mean nBAC} is the chance-normalized mean over all seven datasets, with uncertainty propagated from the per-dataset deviations. Best per column in \textbf{bold}, second best \underline{underlined}.
}
\label{tab:main}
\newcommand{\mcell}[1]{\cellcolor{meanbg}#1}
\begin{center}
\footnotesize
\setlength{\tabcolsep}{8pt}
\begin{tabular}{lrcccc}
\toprule
\textbf{Methods} & \textbf{Size} & \textbf{TUAB} & \textbf{TUEV} & \textbf{ISRUC} & \textbf{Siena} \\
\midrule
EEGNet          & 0.1M   & 0.764 $\pm$ 0.004 & 0.388 $\pm$ 0.014 & 0.715 $\pm$ 0.012 & 0.749 $\pm$ 0.052 \\
EEGConformer    & 0.8M   & 0.776 $\pm$ 0.005 & 0.407 $\pm$ 0.016 & 0.740 $\pm$ 0.013 & 0.756 $\pm$ 0.021 \\
SPaRCNet        & 0.8M & 0.790 $\pm$ 0.002 & 0.416 $\pm$ 0.026 & 0.749 $\pm$ 0.008 & 0.657 $\pm$ 0.038 \\
ContraWR        & 1.6M & 0.775 $\pm$ 0.004 & 0.438 $\pm$ 0.035 & 0.740 $\pm$ 0.013 & 0.655 $\pm$ 0.031 \\
CNN-Transformer & 3.2M   & 0.778 $\pm$ 0.002 & 0.409 $\pm$ 0.016 & 0.736 $\pm$ 0.009 & 0.698 $\pm$ 0.056 \\
ST-Transformer  & 2.6M   & 0.797 $\pm$ 0.002 & 0.398 $\pm$ 0.023 & 0.738 $\pm$ 0.021 & 0.753 $\pm$ 0.038 \\
\midrule
BIOT            & 3.2M & 0.796 $\pm$ 0.006 & 0.528 $\pm$ 0.023 & 0.753 $\pm$ 0.012 & 0.735 $\pm$ 0.067 \\
LaBraM-Base     & 5.8M & 0.814 $\pm$ 0.002 & 0.641 $\pm$ 0.007 & 0.763 $\pm$ 0.010 & 0.708 $\pm$ 0.033 \\
CBraMod         & 4.0M & \underline{0.829} $\pm$ 0.002 & 0.667 $\pm$ 0.011 & \underline{0.787} $\pm$ 0.011 & 0.732 $\pm$ 0.065 \\
REVE-Base       & 69M  & \textbf{0.832} $\pm$ 0.001 & \underline{0.676} $\pm$ 0.023 & 0.782 $\pm$ 0.008 & \underline{0.740} $\pm$ 0.007 \\
\midrule
BrainCodes & \textbf{1.9M} & 0.811 $\pm$ 0.013 & \textbf{0.690} $\pm$ 0.012 & \textbf{0.796} $\pm$ 0.008 & \textbf{0.827} $\pm$ 0.034 \\
\midrule
\textbf{Methods} & \textbf{Size} & \textbf{PhysioNet-MI} & \textbf{BCI IV-2a} & \textbf{EEGMAT} & \mcell{\textbf{Mean nBAC}} \\
\midrule
EEGNet          & 0.1M   & 0.581 $\pm$ 0.013 & 0.448 $\pm$ 0.009 & 0.677 $\pm$ 0.012 & \mcell{0.428 $\pm$ 0.016} \\
EEGConformer    & 0.8M   & 0.605 $\pm$ 0.010 & 0.470 $\pm$ 0.011 & 0.681 $\pm$ 0.012 & \mcell{0.451 $\pm$ 0.008} \\
SPaRCNet        & 0.8M & 0.593 $\pm$ 0.015 & 0.464 $\pm$ 0.012 & 0.688 $\pm$ 0.011 & \mcell{0.428 $\pm$ 0.013} \\
ContraWR        & 1.6M & 0.589 $\pm$ 0.013 & 0.468 $\pm$ 0.013 & 0.663 $\pm$ 0.010 & \mcell{0.418 $\pm$ 0.012} \\
CNN-Transformer & 3.2M   & 0.605 $\pm$ 0.012 & 0.460 $\pm$ 0.011 & 0.678 $\pm$ 0.027 & \mcell{0.432 $\pm$ 0.018} \\
ST-Transformer  & 2.6M   & 0.604 $\pm$ 0.008 & 0.458 $\pm$ 0.015 & 0.663 $\pm$ 0.017 & \mcell{0.446 $\pm$ 0.013} \\
\midrule
BIOT            & 3.2M & 0.615 $\pm$ 0.015 & 0.475 $\pm$ 0.009 & 0.688 $\pm$ 0.019 & \mcell{0.478 $\pm$ 0.021} \\
LaBraM-Base     & 5.8M & 0.617 $\pm$ 0.012 & 0.487 $\pm$ 0.009 & 0.691 $\pm$ 0.013 & \mcell{0.501 $\pm$ 0.011} \\
CBraMod         & 4.0M & 0.642 $\pm$ 0.009 & 0.514 $\pm$ 0.007 & 0.726 $\pm$ 0.013 & \mcell{0.540 $\pm$ 0.019} \\
REVE-Base       & 69M  & \underline{0.648} $\pm$ 0.014 & \textbf{0.640} $\pm$ 0.010 & \underline{0.766} $\pm$ 0.036 & \mcell{\underline{0.581} $\pm$ 0.012} \\
\midrule
BrainCodes & \textbf{1.9M} & \textbf{0.664} $\pm$ 0.006 & \underline{0.639} $\pm$ 0.002 & \textbf{0.772} $\pm$ 0.015 & \mcell{\textbf{0.609} $\pm$ 0.012} \\
\bottomrule
\end{tabular}
\end{center}
\vspace{-5mm}
\end{table}

Table~\ref{tab:main} reports balanced accuracy on the seven standard benchmarks under full fine-tuning, together with the chance-normalized mean over the whole suite. BrainCodes led on five of the seven datasets individually and attained the highest mean nBAC among the foundation models of Table~\ref{tab:main}. Given its lower number of trainable parameters, BrainCodes also yielded the highest mean nBAC per million trainable parameters among foundation models: 0.32 against 0.15 for BIOT, 0.14 for CBraMod, 0.09 for LaBraM and 0.01 for REVE. Notably, while the performance difference on BCI IV-2a does not reach statistical significance, our model trails both REVE-Base and CBraMod on TUAB, which could be attributed to parameter count alone, since TUAB has plenty of available samples, and is larger than our pretraining corpus. On the other hand, the largest margins appear on Siena (+0.087 BAC over REVE) and on mental arithmetic (0.772 against 0.766), the two smallest datasets, and on TUEV (0.690 against 0.676), where the six-way event taxonomy makes chance-normalized gains harder to obtain. Averaged over the suite, BrainCodes recovers 60.9\% of the range between chance and perfect classification against REVE's 58.1\%, using a fraction of its parameters.\vspace{-1mm}

\subsection{Linear Probing Performance}
\vspace{-1mm}

Table~\ref{tab:lp} contrasts our two bottleneck representations against CBraMod and REVE using a frozen encoder. Interestingly, on tasks whose label is a sustained state occupying the whole window, i.e., seizure activity in Siena and cognitive load in mental arithmetic, the pooled feature $b_c$ is ahead by wider margins. On tasks whose label is tied to a transient event at a particular latency, i.e., motor imagery in PhysioNet-MI and BCI IV-2a and event types in TUEV, the flattened feature $\bar{b}_c$ performs better (0.563 against 0.502 on BCI IV-2a, 0.588 against 0.541 on TUEV). Measured against full fine-tuning, the frozen BrainCodes retained 96.2\% on mental arithmetic, 85.1\% on TUEV and 91.2\% on ear-EEG sleep staging. On Siena the linear probe even improved upon fine-tuning, which we attribute to the scarcity of the positive class, on which end-to-end adaptation overfit while the frozen representation acted as a stronger regularizer. The fact that the BrainCodes features remained competitive without any representation learning during downstream adaptation indicates that the transferable structure was learned during pretraining, rather than recovered through fine-tuning.\vspace{-1mm}

\begin{table}[t]
\caption{Linear probe performance on frozen encoders (BAC, mean $\pm$
s.d.\ over 3 seeds). $\mathbf{b}_c$ is the time-averaged bottleneck feature
and $\bar{\mathbf{b}}_c$ the concatenated one, both as defined
in Section~\ref{sec:model}.}
\label{tab:lp}
\begin{center}
\small
\setlength{\tabcolsep}{6.5pt}
\begin{tabular}{lcccc}
\toprule
 & \textbf{CBraMod} & \textbf{REVE-Base} & \multicolumn{2}{c}{\textbf{BrainCodes}} \\
\cmidrule(lr){4-5}
\textbf{Dataset} & Best config & Best config & $\mathbf{b}_c$ & $\bar{\mathbf{b}}_c$ \\
\midrule
EEGMAT (2-class)       & 0.6050 $\pm$ 0.0200 & \underline{0.7400} $\pm$ 0.0730 & \textbf{0.7430} $\pm$ 0.0175 & 0.7097 $\pm$ 0.0225 \\
PhysioNet-MI (4-class) & \underline{0.5310} $\pm$ 0.0150 & \textbf{0.5370} $\pm$ 0.0050 & 0.4137 $\pm$ 0.0076 & 0.4953 $\pm$ 0.0055 \\
BCIC-IV-2a (4-class)   & 0.3760 $\pm$ 0.0060 & \underline{0.5170} $\pm$ 0.0150 & 0.5023 $\pm$ 0.0096 & \textbf{0.5628} $\pm$ 0.0025 \\
TUEV (6-class)         & 0.4820 $\pm$ 0.0370 & \textbf{0.5920} $\pm$ 0.0080 & 0.5406 $\pm$ 0.0052 & \underline{0.5875} $\pm$ 0.0155 \\
Siena (2-class)$^{\ast}$ & 0.7320 $\pm$ 0.0660 & 0.7400 $\pm$ 0.0070 & \textbf{0.8747} $\pm$ 0.0005 & \underline{0.7830} $\pm$ 0.0340 \\
ISRUC (5-class)        & 0.4300 $\pm$ 0.0430 & 0.6970 $\pm$ 0.0110 & \underline{0.7147} $\pm$ 0.0005 & \textbf{0.7231} $\pm$ 0.0007 \\
EESM23 (6-class)        & 0.4002 $\pm$ 0.0110 & 0.3348 $\pm$ 0.0235 & \textbf{0.4791} $\pm$ 0.0041 & \underline{0.4412} $\pm$ 0.0099 \\
\midrule
\textbf{Mean nBAC} & 0.3016 $\pm$ 0.0310 & 0.4895 $\pm$ 0.0298 &
\multicolumn{2}{c}{Best config: \textbf{0.5518} $\pm$ 0.0073} \\
\bottomrule
\end{tabular}
\end{center}
 \small $^{\ast}$No published linear-probe baseline exists for Siena, so we compare against full fine-tuning performance.
 \vspace{-4mm}
\end{table}

\subsection{Application on Wearable Devices}
\label{sec:wearable}
\vspace{-1mm}

To our knowledge, we provide the first benchmark evaluation of EEG foundation models on consumer wearable devices. Specifically, we evaluate transfer to two forms of consumer hardware: the four-channel Muse headband (TP9, AF7, AF8, TP10) and EESM23 sleep staging from in-ear electrodes~\citep{bjarke2025ear}. All models were evaluated on identical splits, i.e., four-fold cross-validation over participants for the two Muse tasks and the fixed participant split of \citet{tfmtokenizer} for EESM23, and the results are shared in Table~\ref{tab:wearable}.
 
On the motor imagery task, BrainCodes reaches 0.5889 BAC against 0.5083 for the best baseline, whereas Muse-mental is a tie for all frontier models, as every model's standard deviation exceeds the between-model spread. On EESM23, BrainCodes reaches 0.5254 against 0.4176 for REVE, i.e., a clear head over the strongest baseline. Ear-EEG is the setting furthest from the pretraining montage, yet a frozen linear probe still reached 0.4791, i.e., 91\% of the full fine-tuning performance.\vspace{-1mm}

\begin{table}[t]
\caption{Balanced accuracy on low-resource, consumer wearables. Muse results are means over 4-fold cross-validation; EESM23 over 3 seeds on the test set used in~\citep{tfmtokenizer}.}
\label{tab:wearable}
\begin{center}
\small
\setlength{\tabcolsep}{6.6pt}
\begin{tabular}{lrccc>{\columncolor{meanbg}}c}
\toprule
\textbf{Model} & \textbf{Size} & \textbf{EESM23} & \textbf{Muse-Motor} & \textbf{Muse-Mental} & \textbf{Mean nBAC} \\
\midrule
LaBraM-Base   & 5.8M & 0.4109 $\pm$ 0.0087 & 0.4720 $\pm$ 0.0048 & \textbf{0.8245} $\pm$ 0.1446 & 0.4126 $\pm$ 0.0724 \\
CBraMod       & 4.0M & 0.3993 $\pm$ 0.0163 & \underline{0.5083} $\pm$ 0.0267 & 0.8079 $\pm$ 0.1812 & \underline{0.4178} $\pm$ 0.0918 \\
CSBrain       & 8.9M & 0.3663 $\pm$ 0.0261 & 0.4452 $\pm$ 0.0349 & 0.7081 $\pm$ 0.2143 & 0.3232 $\pm$ 0.1091 \\
REVE-Base     & 69M  & \underline{0.4176} $\pm$ 0.0224 & 0.4485 $\pm$ 0.0240 & 0.7165 $\pm$ 0.2177 & 0.3495 $\pm$ 0.1099 \\
\midrule
BrainCodes    & \textbf{1.9M} & \textbf{0.5254} $\pm$ 0.0251 & \textbf{0.5889} $\pm$ 0.0318 & \underline{0.8163} $\pm$ 0.0957 & \textbf{0.5128} $\pm$ 0.0514 \\
\bottomrule
\end{tabular}
\end{center}
\vspace{-5mm}
\end{table}

\section{Grounding Analysis}
\label{sec:suite}

\begin{figure}
    \centering
    \includegraphics[width=0.98\linewidth, height=273pt]{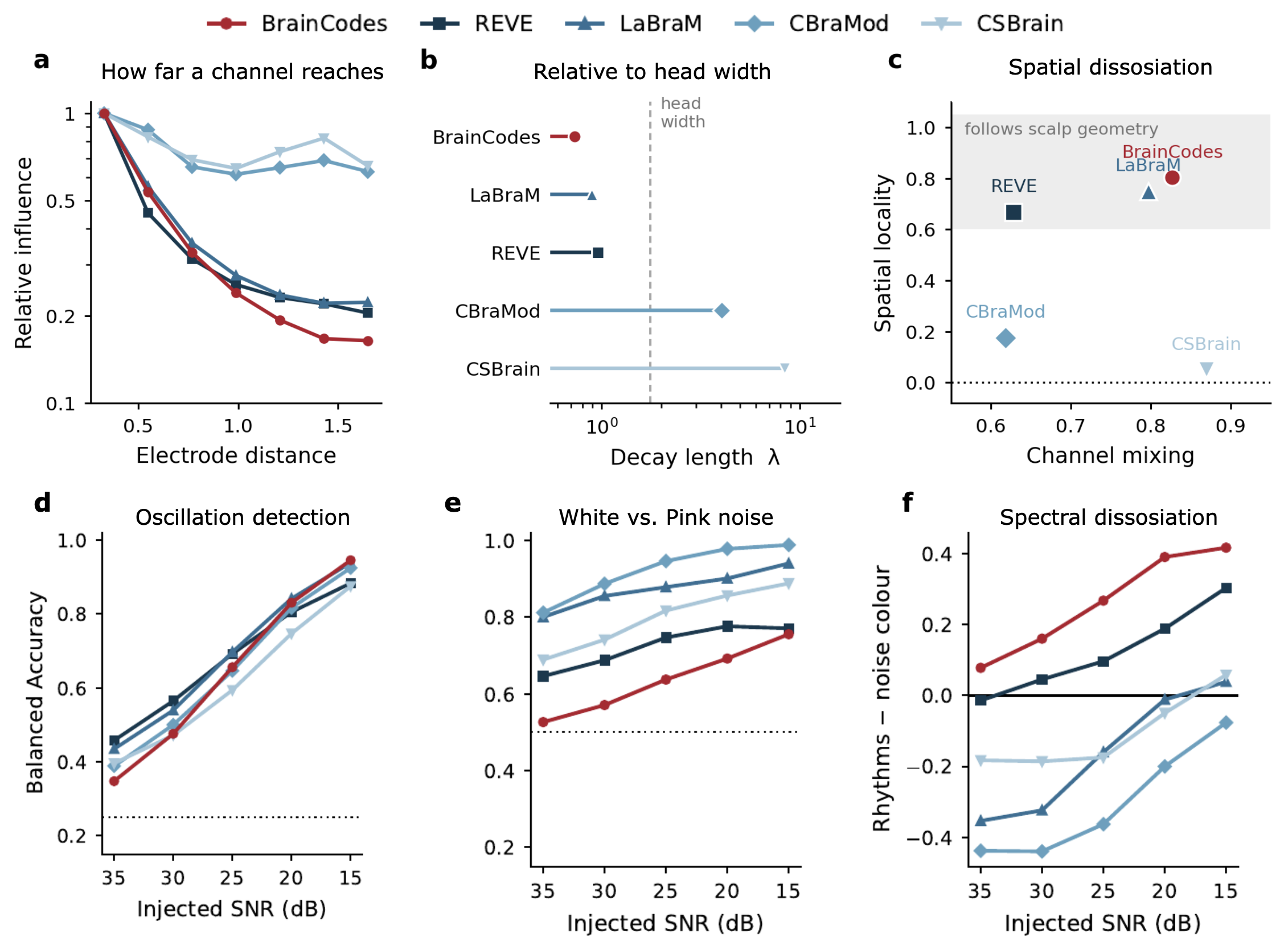}
    \vspace{-4mm}
    \caption{\textbf{Grounding analysis of BrainCodes.} \textbf{(a--c)} \textit{Spatial} organization: influence of one electrode on another against inter-electrode distance, the fitted decay length $\lambda$ relative to head width, and channel mixing against spatial locality. \textbf{(d--f)} \textit{Spectral} selectivity under four-way injected rhythm identification, white-versus-pink noise control, and their difference as a preference index.}
    \label{fig:grounding}
    \vspace{-5mm}
\end{figure}

The downstream results establish that BrainCodes is transferable and parameter-efficient, but predictive performance alone cannot validate the motivation of its objective, i.e., whether the learned representation preserves oscillatory and spatial structure that typical reconstruction may underweight. The following analyses probe whether channel interactions follow electrode geometry and whether the representation preferentially exposes narrowband rhythms or merely broadband variation.\vspace{-1mm}

\subsection{Retained Spatial Dynamics}
\label{sec:spatial}

A desirable property of a foundation model for EEG is the ability to reflect spatial dependencies between channels that follow the scalp geometry. We test this explicitly through two properties: \emph{mixing}, whether perturbing one electrode changes the representation of other channels at all (Figure~\ref{fig:grounding}a), and \emph{locality}, the rank correlation between the size of that change and inter-electrode distance (Figure~\ref{fig:grounding}b), which dissociate in practice (Figure~\ref{fig:grounding}c). CSBrain, BrainCodes and LaBraM propagate induced perturbations most widely across the montage, while REVE and CBraMod confine them more narrowly, as shown in the 2D channel matrices.

Locality, however, is largely independent of mixing. CBraMod and CSBrain inherit no grounded spatial structure, so their mixing spreads indiscriminately across the scalp rather than decaying with distance. BrainCodes and LaBraM respect scalp geometry the most, which we attribute to their explicit channel positional encoding. BrainCodes also has the shortest influence decay length, and the only one comfortably within the size of the scalp ($\lambda = 0.73$, against $0.90$ for LaBraM and $0.96$ for REVE, while CBraMod and CSBrain reach $4.05$ and $4.53$). Taken together, it emerges as the most spatially grounded model along both axes of this analysis, which supports the proposed positional encoding based on RoPE embeddings of learnable scalp directions.\vspace{-2mm}

\subsection{Retained Spectral Dynamics}
\vspace{-1mm}

Because broadband aperiodic activity contributes substantial signal power, squared-error reconstruction may prioritize it over lower-power oscillatory structure~\citep{donoghue2020parameterizing, kommineni2026aperiodic}. We tested this effect by injecting controlled perturbations of two kinds into 512 held-out test segments of PhysioNet-MI and identifying them through linear probing. The first was a narrowband burst drawn from one of four canonical rhythms ($\delta$, $\mu$, $\beta$, spindle), i.e., a four-way discrimination that required the representation to encode oscillatory information of physiological interest. The second was a noise control task between white and pink noise, which only required the slope of the broadband spectrum. Sweeping the injected SNR from 35 to 15\,dB traces how each property becomes recoverable as the perturbation strengthens (Figure~\ref{fig:grounding}\,d,e).

Rhythm identification was comparable across models, with BrainCodes reaching the highest accuracy at the strongest injection (0.95 against 0.88 for REVE). On noise color though, BrainCodes was the least accurate model at every SNR, rising only from 0.53 to 0.76, while CBraMod reached 0.99. Taking the difference between the two curves in chance-normalized balanced accuracy, which accounted for their chance levels of 0.25 and 0.5 (Figure~\ref{fig:grounding}f), gives a single interpretable index of what each latent space preferentially encodes. BrainCodes was the only model for which this index is positive across the entire sweep, followed by REVE. Interestingly, these models also performed the best across the benchmarks of Section~\ref{sec:results}. We read this as evidence and motivation for the use of the auxiliary objective, and Appendix~\ref{app:spectral} verifies that this sensitivity to narrowband oscillatory activity extends to naturally occurring rhythms.\vspace{-2mm}

\section{Conclusion}
\label{sec:conclusion}
\vspace{-2mm}

We introduced \textbf{BrainCodes}, a lightweight EEG foundation model combining discrete neural token prediction with montage-aware spatial encoding and oscillatory supervision. Across 14 benchmarks, BrainCodes matched or outperformed larger models, transferred effectively to sparse consumer montages, and preserved measurable spatial and spectral structure. Our findings show that grounded objectives can support EEG representation learning at substantially smaller scale. Validation across more diverse populations and grounded probes would further strengthen this scope.


\subsection*{AI Use Statement}


In this work we used generative AI tools as a general-purpose writing and coding assistant. Specifically, we used them to polish grammar and clarity, to draft and refine \LaTeX{} tables and figure captions from experimental result files, and to assist with plotting utilities, execution of experiments and debugging training scripts. We did not use generative AI tools for research ideation, for the design of the method or its objectives, for the design of experiments, for the analysis or interpretation of results, or for retrieving related work or references. All scientific claims, experimental designs, and interpretations in this paper originate from the authors. Use of generative AI as a subject of study, as a component of the proposed method, or for the generation of any data used in training or evaluation is not applicable to this work. The authors have reviewed all AI-assisted work and therefore take full responsibility for the final content of this study.

\subsection*{Ethics Statement}

This work is built solely on publicly released, de-identified EEG datasets and involves no new data collection, no human-subjects interaction, and no attempt at re-identification. The pretraining corpus is the Healthy Brain Network EEG release~\citep{alexander2017open}, collected by the Child Mind Institute under its own institutional review board approval, and distributed in de-identified form under the terms of its data use agreement, which we have followed. We note that this corpus consists of children and adolescents, a population for whom the downstream consequences of representation learning are less well characterized than for adults; we therefore make no claims about clinical or diagnostic validity in this or any other population based on our model's inference. All downstream datasets are public benchmarks used in their standard released form, and our participant-level splits ensure that no individual contributes to both a training and an evaluation set.

We draw attention to two risks. First, several of the evaluation tasks are clinical in nature and the accuracies reported here are benchmark results on curated retrospective datasets, not evidence of fitness for clinical deployment; using BrainCodes to inform patient care without prospective validation and regulatory review would be inappropriate. Second, EEG is a biometric signal from which attributes beyond the labeled task may be recoverable, so representations learned from it should be treated as identifiable data and handled accordingly. The authors declare no competing interests that could have influenced the design, execution, or reporting of this work.

\subsection*{Reproducibility Statement}

We have taken the following steps to support reproduction of our results. The pretraining corpus, preprocessing pipeline, montage reduction, and windowing and tokenization procedure are specified in Section~\ref{sec:setup}, including filter settings, sampling rate, codebook configuration, and code-frame rate. The architecture, positional encoding, bottleneck representations, and both pretraining objectives are given in Section~\ref{sec:methods}, and the optimization settings, hardware, and training schedule are given in Section~\ref{sec:setup}. All downstream datasets are public; their sizes, montages, class counts, and the splitting protocol used for each are listed in Table~\ref{tab:datasets}, and the adaptation protocols, metric definition, and seed counts are given in Section~\ref{sec:setup}. Every result in the main text is reported as a mean and standard deviation over the seeds or folds stated in the corresponding caption. Anonymized source code for pretraining, fine-tuning, and all grounding probes, together with the configuration files needed to reproduce each table and figure, is included in the supplementary materials.

\bibliography{iclr2027_conference}

@article{eegnet,
  author  = {Lawhern, Vernon J. and Solon, Amelia J. and Waytowich, Nicholas R. and Gordon, Stephen M. and Hung, Chou P. and Lance, Brent J.},
  title   = {{EEGNet}: A Compact Convolutional Network for {EEG}-based Brain--Computer Interfaces},
  journal = {Journal of Neural Engineering},
  volume  = {15},
  number  = {5},
  pages   = {056013},
  year    = {2018},
  doi     = {10.1088/1741-2552/aace8c},
  eprint  = {1611.08024},
  archivePrefix = {arXiv},
}

@article{sttransformer,
  author  = {Song, Yonghao and Jia, Xueyu and Yang, Lie and Xie, Longhan},
  title   = {Transformer-based Spatial-Temporal Feature Learning for {EEG} Decoding},
  journal = {arXiv preprint arXiv:2106.11170},
  year    = {2021},
  eprint  = {2106.11170},
  archivePrefix = {arXiv},
}

@inproceedings{eegpt,
  author    = {Wang, Guangyu and Liu, Wenchao and He, Yuhong and Xu, Cong and Ma, Lin and Li, Haifeng},
  title     = {{EEGPT}: Pretrained Transformer for Universal and Reliable Representation of {EEG} Signals},
  booktitle = {Advances in Neural Information Processing Systems (NeurIPS)},
  year      = {2024},
}

@inproceedings{cbramod,
  author    = {Wang, Jiquan and Zhao, Sha and Luo, Zhiling and Zhou, Yangxuan and Jiang, Haiteng and Li, Shijian and Li, Tao and Pan, Gang},
  title     = {{CBraMod}: A Criss-Cross Brain Foundation Model for {EEG} Decoding},
  booktitle = {International Conference on Learning Representations (ICLR)},
  year      = {2025},
}

@inproceedings{csbrain,
  author    = {Zhou, Yuchen and Wu, Jiamin and Ren, Zichen and Yao, Zhouheng and Lu, Weiheng and Peng, Kunyu and Zheng, Qihao and Song, Chunfeng and Ouyang, Wanli and Gou, Chao},
  title     = {{CSBrain}: A Cross-scale Spatiotemporal Brain Foundation Model for {EEG} Decoding},
  booktitle = {Advances in Neural Information Processing Systems (NeurIPS)},
  year      = {2025},
  eprint    = {2506.23075},
  archivePrefix = {arXiv},
}

@inproceedings{reve,
  author    = {El Ouahidi, Yassine and Lys, Jonathan and Th{\"o}lke, Philipp and Farrugia, Nicolas and Pasdeloup, Bastien and Gripon, Vincent and Jerbi, Karim and Lioi, Giulia},
  title     = {{REVE}: A Foundation Model for {EEG} -- Adapting to Any Setup with Large-Scale Pretraining on 25,000 Subjects},
  booktitle = {Advances in Neural Information Processing Systems (NeurIPS)},
  year      = {2025},
  eprint    = {2510.21585},
  archivePrefix = {arXiv},
}

@inproceedings{biot,
  author    = {Yang, Chaoqi and Westover, M. Brandon and Sun, Jimeng},
  title     = {{BIOT}: Cross-data Biosignal Learning in the Wild},
  booktitle = {Advances in Neural Information Processing Systems (NeurIPS)},
  year      = {2023},
  eprint    = {2305.10351},
  archivePrefix = {arXiv},
}

@inproceedings{labram,
  author    = {Jiang, Wei-Bang and Zhao, Li-Ming and Lu, Bao-Liang},
  title     = {Large Brain Model for Learning Generic Representations with Tremendous {EEG} Data in {BCI}},
  booktitle = {International Conference on Learning Representations (ICLR)},
  year      = {2024},
  eprint    = {2405.18765},
  archivePrefix = {arXiv},
}

@article{neurorvq,
  author  = {Barmpas, Konstantinos and Lee, Na and Chalatsis, Dimitrios and Raftery, William and Panagakis, Yannis and Adamos, Dimitrios A. and Laskaris, Nikolaos and Koliousis, Alexandros and Farina, Dario and Zafeiriou, Stefanos},
  title   = {{NeuroRVQ}: Multi-Scale Biosignal Tokenization for Generative Foundation Models},
  journal = {arXiv preprint arXiv:2510.13068},
  year    = {2025},
  eprint  = {2510.13068},
  archivePrefix = {arXiv},
}

@article{brainomni,
  author  = {Xiao, Qinfan and Cui, Ziyun and Zhang, Chi and Chen, Siqi and Wu, Wen and Thwaites, Andrew and Woolgar, Alexandra and Zhou, Bowen and Zhang, Chao},
  title   = {{BrainOmni}: A Brain Foundation Model for Unified {EEG} and {MEG} Signals},
  journal = {arXiv preprint arXiv:2505.18185},
  year    = {2025},
  eprint  = {2505.18185},
  archivePrefix = {arXiv},
}

@article{tfmtokenizer,
  author  = {Pradeepkumar, Jathurshan and Piao, Xihao and Chen, Zheng and Sun, Jimeng},
  title   = {Tokenizing Single-Channel {EEG} with Time-Frequency Motif Learning},
  journal = {arXiv preprint arXiv:2502.16060},
  year    = {2025},
  eprint  = {2502.16060},
  archivePrefix = {arXiv},
}

@article{biocodec,
  author  = {Avramidis, Kleanthis and Feng, Tiantian and Jeong, Woojae and Lee, Jihwan and Cui, Wenhui and Leahy, Richard M. and Narayanan, Shrikanth},
  title   = {Neural Codecs as Biosignal Tokenizers},
  journal = {arXiv:2510.09095},
  year    = {2025},
  eprint  = {2510.09095},
  archivePrefix = {arXiv},
}

@article{he2014,
  author  = {He, Biyu J.},
  title   = {Scale-free Brain Activity: Past, Present, and Future},
  journal = {Trends in Cognitive Sciences},
  volume  = {18},
  number  = {9},
  pages   = {480--487},
  year    = {2014},
  doi     = {10.1016/j.tics.2014.04.003},
}

@article{fooof,
  author  = {Donoghue, Thomas and Haller, Matar and Peterson, Erik J. and Varma, Paroma and Sebastian, Priyadarshini and Gao, Richard and Noto, Torben and Lara, Antonio H. and Wallis, Joni D. and Knight, Robert T. and Shestyuk, Avgusta and Voytek, Bradley},
  title   = {Parameterizing Neural Power Spectra into Periodic and Aperiodic Components},
  journal = {Nature Neuroscience},
  volume  = {23},
  number  = {12},
  pages   = {1655--1665},
  year    = {2020},
  doi     = {10.1038/s41593-020-00744-x},
}

@book{nunez,
  author    = {Nunez, Paul L. and Srinivasan, Ramesh},
  title     = {Electric Fields of the Brain: The Neurophysics of {EEG}},
  edition   = {2nd},
  publisher = {Oxford University Press},
  year      = {2006},
}

@article{perrin1989,
  author  = {Perrin, F. and Pernier, J. and Bertrand, O. and Echallier, J. F.},
  title   = {Spherical Splines for Scalp Potential and Current Density Mapping},
  journal = {Electroencephalography and Clinical Neurophysiology},
  volume  = {72},
  number  = {2},
  pages   = {184--187},
  year    = {1989},
  doi     = {10.1016/0013-4694(89)90180-6},
}

@article{obeid2016temple,
  title={The temple university hospital EEG data corpus},
  author={Obeid, Iyad and Picone, Joseph},
  journal={Frontiers in neuroscience},
  volume={10},
  pages={196},
  year={2016},
  publisher={Frontiers Media SA}
}

@article{detti2020eeg,
  title={EEG synchronization analysis for seizure prediction: A study on data of noninvasive recordings},
  author={Detti, Paolo and Vatti, Giampaolo and Zabalo Manrique de Lara, Garazi},
  journal={Processes},
  volume={8},
  number={7},
  pages={846},
  year={2020},
  publisher={MDPI}
}

@article{alexander2017open,
  title={An open resource for transdiagnostic research in pediatric mental health and learning disorders},
  author={Alexander, Lindsay M and Escalera, Jasmine and Ai, Lei and Andreotti, Charissa and Febre, Karina and Mangone, Alexander and Vega-Potler, Natan and Langer, Nicolas and Alexander, Alexis and Kovacs, Meagan and others},
  journal={Scientific data},
  volume={4},
  number={1},
  pages={170181},
  year={2017},
  publisher={Nature Publishing Group}
}

@article{kemp2000analysis,
  title={Analysis of a sleep-dependent neuronal feedback loop: the slow-wave microcontinuity of the EEG},
  author={Kemp, Bob and Zwinderman, Aeilko H and Tuk, Bert and Kamphuisen, Hilbert AC and Oberye, Josefien JL},
  journal={IEEE Transactions on Biomedical Engineering},
  volume={47},
  number={9},
  pages={1185--1194},
  year={2000},
  publisher={IEEE}
}

@article{bjarke2025ear,
  title={Ear-EEG sleep monitoring data sets},
  author={Bjarke Mikkelsen, Kaare and Rezai Tabar, Yousef and R{\ae}vsb{\ae}k Birch, Laura and Lind Kappel, Simon and Bech Christensen, Christian and Dalskov Mosgaard, Lars and Otto, Marit and Christian Hemmsen, Martin and Lind Rank, Mike and Kidmose, Preben},
  journal={Scientific Data},
  volume={12},
  number={1},
  pages={301},
  year={2025},
  publisher={Nature Publishing Group UK London}
}

@article{su2024roformer,
  title={Roformer: Enhanced transformer with rotary position embedding},
  author={Su, Jianlin and Ahmed, Murtadha and Lu, Yu and Pan, Shengfeng and Bo, Wen and Liu, Yunfeng},
  journal={Neurocomputing},
  volume={568},
  pages={127063},
  year={2024},
  publisher={Elsevier}
}

@inproceedings{heo2024rotary,
  title={Rotary position embedding for vision transformer},
  author={Heo, Byeongho and Park, Song and Han, Dongyoon and Yun, Sangdoo},
  booktitle={European Conference on Computer Vision},
  pages={289--305},
  year={2024},
  organization={Springer}
}

@article{khalighi2016isruc,
  title={ISRUC-Sleep: A comprehensive public dataset for sleep researchers},
  author={Khalighi, Sirvan and Sousa, Teresa and Santos, Jos{\'e} Moutinho and Nunes, Urbano},
  journal={Computer methods and programs in biomedicine},
  volume={124},
  pages={180--192},
  year={2016},
  publisher={Elsevier}
}

@article{schalk2004bci2000,
  title={BCI2000: a general-purpose brain-computer interface (BCI) system},
  author={Schalk, Gerwin and McFarland, Dennis J and Hinterberger, Thilo and Birbaumer, Niels and Wolpaw, Jonathan R},
  journal={IEEE Transactions on biomedical engineering},
  volume={51},
  number={6},
  pages={1034--1043},
  year={2004},
  publisher={IEEE}
}

@article{brunner2008bci,
  title={BCI Competition 2008--Graz data set A},
  author={Brunner, Clemens and Leeb, Robert and M{\"u}ller-Putz, Gernot and Schl{\"o}gl, Alois and Pfurtscheller, Gert},
  journal={Institute for knowledge discovery (laboratory of brain-computer interfaces), Graz University of Technology},
  volume={16},
  number={1-6},
  pages={34},
  year={2008}
}

@article{zyma2019electroencephalograms,
  title={Electroencephalograms during mental arithmetic task performance},
  author={Zyma, Igor and Tukaev, Sergii and Seleznov, Ivan and Kiyono, Ken and Popov, Anton and Chernykh, Mariia and Shpenkov, Oleksii},
  journal={Data},
  volume={4},
  number={1},
  pages={14},
  year={2019},
  publisher={MDPI}
}

@article{schirrmeister2017deep,
  title={Deep learning with convolutional neural networks for EEG decoding and visualization},
  author={Schirrmeister, Robin Tibor and Springenberg, Jost Tobias and Fiederer, Lukas Dominique Josef and Glasstetter, Martin and Eggensperger, Katharina and Tangermann, Michael and Hutter, Frank and Burgard, Wolfram and Ball, Tonio},
  journal={Human brain mapping},
  volume={38},
  number={11},
  pages={5391--5420},
  year={2017},
  publisher={Wiley Online Library}
}

@article{lee2019eeg,
  title={EEG dataset and OpenBMI toolbox for three BCI paradigms: An investigation into BCI illiteracy},
  author={Lee, Min-Ho and Kwon, O-Yeon and Kim, Yong-Jeong and Kim, Hong-Kyung and Lee, Young-Eun and Williamson, John and Fazli, Siamac and Lee, Seong-Whan},
  journal={GigaScience},
  volume={8},
  number={5},
  pages={giz002},
  year={2019},
  publisher={Oxford University Press}
}

@article{pavlov2022pupillometry,
  title={Pupillometry and electroencephalography in the digit span task},
  author={Pavlov, Yuri G and Kasanov, Dauren and Kosachenko, Alexandra I and Kotyusov, Alexander I and Busch, Niko A},
  journal={Scientific data},
  volume={9},
  number={1},
  pages={325},
  year={2022},
  publisher={Nature Publishing Group UK London}
}

@article{lawhern_eegnet_2018,
    title = {{EEGNet}: a compact convolutional neural network for {EEG}-based brain–computer interfaces},
    volume = {15},
    issn = {1741-2560, 1741-2552},
    shorttitle = {{EEGNet}},
    url = {https://iopscience.iop.org/article/10.1088/1741-2552/aace8c},
    doi = {10.1088/1741-2552/aace8c},
    language = {en},
    number = {5},
    urldate = {2024-10-28},
    journal = {Journal of Neural Engineering},
    author = {Lawhern, Vernon J and Solon, Amelia J and Waytowich, Nicholas R and Gordon, Stephen M and Hung, Chou P and Lance, Brent J},
    month = oct,
    year = {2018},
    pages = {056013},
}

@article{jing2023development,
  title={Development of expert-level classification of seizures and rhythmic and periodic patterns during EEG interpretation},
  author={Jing, Jin and Ge, Wendong and Hong, Shenda and Fernandes, Marta Bento and Lin, Zhen and Yang, Chaoqi and An, Sungtae and Struck, Aaron F and Herlopian, Aline and Karakis, Ioannis and others},
  journal={Neurology},
  volume={100},
  number={17},
  pages={e1750--e1762},
  year={2023},
  publisher={Lippincott Williams \& Wilkins Hagerstown, MD}
}

@article{yang2023self,
  title={Self-supervised electroencephalogram representation learning for automatic sleep staging: model development and evaluation study},
  author={Yang, Chaoqi and Xiao, Cao and Westover, M Brandon and Sun, Jimeng},
  journal={JMIR AI},
  volume={2},
  number={1},
  pages={e46769},
  year={2023},
  publisher={JMIR Publications Inc., Toronto, Canada}
}

@article{krigolson2017choosing,
  title={Choosing MUSE: Validation of a low-cost, portable EEG system for ERP research},
  author={Krigolson, Olave E and Williams, Chad C and Norton, Angela and Hassall, Cameron D and Colino, Francisco L},
  journal={Frontiers in neuroscience},
  volume={11},
  pages={109},
  year={2017},
  publisher={Frontiers Media SA}
}

@inproceedings{bird2018study,
  title={A study on mental state classification using eeg-based brain-machine interface},
  author={Bird, Jordan J and Manso, Luis J and Ribeiro, Eduardo P and Ekart, Aniko and Faria, Diego R},
  booktitle={2018 international conference on intelligent systems (IS)},
  pages={795--800},
  year={2018},
  organization={IEEE}
}

@article{kommineni2026aperiodic,
  title={Aperiodic and Low-Frequency Spectral Bias in Reconstruction based EEG Foundation Models},
  author={Kommineni, Aditya and Zhou, Emily and Avramidis, Kleanthis and Segaard, Simon Bock and M{\"u}nster, Jeppe Roden and Hansen, Andreas Peter Juhl and Medani, Takfarinas and Feng, Tiantian and Leahy, Richard and Narayanan, Shrikanth},
  journal={arXiv preprint arXiv:2605.26434},
  year={2026}
}

@article{pollard_physionet_2026, title = {{PhysioNet} as a global platform for biomedical research}, volume = {1}, issn = {3005-0693}, url = {https://doi.org/10.1038/s44360-026-00096-z}, doi = {10.1038/s44360-026-00096-z}, number = {8}, journal = {Nature Health}, author = {Pollard, Tom and Moody, Benjamin E. and Lehman, Li-wei H. and Gow, Brian J. and Fernandes, Chrystinne and Xie, Chen and Johnson, Alistair and Mark, Roger G. and Heldt, Thomas}, month = aug, year = {2026}, pages = {792--795},}

@inproceedings{neurogpt,
  title={Neuro-gpt: Towards a foundation model for eeg},
  author={Cui, Wenhui and Jeong, Woojae and Th{\"o}lke, Philipp and Medani, Takfarinas and Jerbi, Karim and Joshi, Anand A and Leahy, Richard M},
  booktitle={2024 IEEE International Symposium on Biomedical Imaging (ISBI)},
  pages={1--5},
  year={2024},
  organization={IEEE}
}

@article{song2022eeg,
  title={EEG conformer: Convolutional transformer for EEG decoding and visualization},
  author={Song, Yonghao and Zheng, Qingqing and Liu, Bingchuan and Gao, Xiaorong},
  journal={IEEE Transactions on Neural Systems and Rehabilitation Engineering},
  volume={31},
  pages={710--719},
  year={2022},
  publisher={IEEE}
}

@inproceedings{peh2022transformer,
  title={Transformer convolutional neural networks for automated artifact detection in scalp EEG},
  author={Peh, Wei Yan and Yao, Yuanyuan and Dauwels, Justin},
  booktitle={2022 44th Annual International Conference of the IEEE Engineering in Medicine \& Biology Society (EMBC)},
  pages={3599--3602},
  year={2022},
  organization={IEEE}
}

@article{song2021transformer,
  title={Transformer-based spatial-temporal feature learning for EEG decoding},
  author={Song, Yonghao and Jia, Xueyu and Yang, Lie and Xie, Longhan},
  journal={arXiv preprint arXiv:2106.11170},
  year={2021}
}

@article{gramfort2013meg,
  title={MEG and EEG data analysis with MNE-Python},
  author={Gramfort, Alexandre and Luessi, Martin and Larson, Eric and Engemann, Denis A and Strohmeier, Daniel and Brodbeck, Christian and Goj, Roman and Jas, Mainak and Brooks, Teon and Parkkonen, Lauri and others},
  journal={Frontiers in Neuroinformatics},
  volume={7},
  pages={267},
  year={2013},
  publisher={Frontiers Media SA}
}

@article{paszke2019pytorch,
  title={Pytorch: An imperative style, high-performance deep learning library},
  author={Paszke, Adam and Gross, Sam and Massa, Francisco and Lerer, Adam and Bradbury, James and Chanan, Gregory and Killeen, Trevor and Lin, Zeming and Gimelshein, Natalia and Antiga, Luca and others},
  journal={Advances in neural information processing systems},
  volume={32},
  year={2019}
}

@article{perslev2021u,
  title={U-Sleep: resilient high-frequency sleep staging},
  author={Perslev, Mathias and Darkner, Sune and Kempfner, Lykke and Nikolic, Miki and Jennum, Poul J{\o}rgen and Igel, Christian},
  journal={NPJ digital medicine},
  volume={4},
  number={1},
  pages={72},
  year={2021},
  publisher={Nature Publishing Group UK London}
}

@article{supratak2017deepsleepnet,
  title={DeepSleepNet: A model for automatic sleep stage scoring based on raw single-channel EEG},
  author={Supratak, Akara and Dong, Hao and Wu, Chao and Guo, Yike},
  journal={IEEE transactions on neural systems and rehabilitation engineering},
  volume={25},
  number={11},
  pages={1998--2008},
  year={2017},
  publisher={IEEE}
}

@article{donoghue2020parameterizing,
  title={Parameterizing neural power spectra into periodic and aperiodic components},
  author={Donoghue, Thomas and Haller, Matar and Peterson, Erik J and Varma, Paroma and Sebastian, Priyadarshini and Gao, Richard and Noto, Torben and Lara, Antonio H and Wallis, Joni D and Knight, Robert T and others},
  journal={Nature neuroscience},
  volume={23},
  number={12},
  pages={1655--1665},
  year={2020},
  publisher={Nature Publishing Group US New York}
}

@article{bomatter2026limited,
  title={Is Limited Participant Diversity Impeding EEG-based Machine Learning?},
  author={Bomatter, Philipp and Gouk, Henry},
  journal={Advances in Neural Information Processing Systems},
  volume={38},
  pages={117447--117474},
  year={2026}
}

@misc{elkaffas2024musemotor,
  author       = {Elkaffas, Muhammad Atef},
  title        = {{EEG Muse2 Motor Imagery Brain Electrical Activity}},
  year         = {2024},
  howpublished = {Kaggle dataset},
  url          = {https://www.kaggle.com/datasets/muhammadatefelkaffas/eeg-muse2-motor-imagery-brain-electrical-activity},
  note         = {Accessed: 2026-09-26}
}
\bibliographystyle{iclr2027_conference}

\newpage
\appendix

\section{Related Work}

\textbf{Self-supervised Learning for EEG.} Most EEG foundation models learn by reconstructing masked segments of the input. LaBraM first trained a vector-quantized tokenizer to encode the Fourier amplitude and phase of EEG patches, then pretrained a transformer on approximately 2,500 hours of recordings to predict the indices of masked patches~\citep{labram}. CBraMod instead reconstructed masked waveform patches using a criss-cross transformer with separate spatial and temporal attention, pretrained on more than 9,000 hours from the TUEG corpus~\citep{cbramod}. REVE scaled masked autoencoding to more than 60,000 hours from 25,000 participants across 92 datasets, with model sizes ranging from 12M to 408M parameters~\citep{reve}. BIOT represented each channel independently as a sequence of segment embeddings processed by a linear-attention transformer, allowing it to train across datasets with different channel sets~\citep{biot}. CSBrain, in turn, combined cross-scale spatiotemporal tokenization with structured sparse attention~\citep{csbrain}. Despite their architectural differences, these models optimize reconstruction targets whose contribution varies with spectral power. Reconstruction-based EEG models consequently tend to emphasize aperiodic and low-frequency content~\citep{kommineni2026aperiodic}, motivating objectives whose weighting is less directly tied to signal amplitude.

\textbf{Discrete Tokenization of Biosignals.} Neural audio codecs and vector quantization have also motivated discrete representations of physiological signals. BioCodec applied residual vector quantization to single-channel biosignals and showed that its codes support downstream decoding at a fraction of the original bitrate~\citep{biocodec}. NeuroRVQ used a multi-scale residual quantizer within a generative EEG foundation model~\citep{neurorvq}, while TFM-Tokenizer learned a vocabulary of time--frequency motifs from single-channel EEG and evaluated it on tasks including in-ear sleep staging~\citep{tfmtokenizer}. These tokenizers were evaluated primarily through downstream accuracy, providing actionable insights on using vector quantization as a denoising method for EEG representations.

\textbf{Spatial Encoding of Electrodes.} Transfer across heterogeneous montages requires electrode position to be represented in a form that remains meaningful across layouts. LaBraM learned a separate spatial embedding for each electrode name~\citep{labram}, while CBraMod generated positional encodings from the input through a convolutional module~\citep{cbramod}. REVE jointly encoded electrode coordinates and time using a four-dimensional Fourier positional encoding~\citep{reve}. Rotary position embeddings offer a complementary approach: they encode position through rotations of query and key pairs, making the resulting attention logits depend on relative offset~\citep{su2024roformer}. Their mixed-axis extension learns the rotation frequency and orientation of each pair jointly over two-dimensional image coordinates~\citep{heo2024rotary}. Extending this formulation to three-dimensional electrode coordinates makes the positional component of channel attention depend only on inter-electrode displacement, the geometric quantity most directly related to volume conduction at the scalp~\citep{nunez}.

\textbf{EEG on Consumer Hardware.} Low-cost headbands and ear-EEG systems have been validated for event-related potential research~\citep{krigolson2017choosing} and sleep monitoring~\citep{bjarke2025ear}, and mental states have been classified from the four electrodes of the Muse headband~\citep{bird2018study}. These devices nevertheless present a substantial transfer challenge: their montages contain only four electrodes, often at positions that are sparse or absent in the recordings used for pretraining. Existing EEG foundation models have been evaluated almost entirely on clinical and laboratory montages. Even in-ear sleep staging has been studied mainly by using BIOT and LaBraM as front ends to TFM-Tokenizer~\citep{tfmtokenizer}. Direct transfer of pretrained EEG representations to consumer hardware therefore remains largely unexplored.

\section{Model Performance: \textsc{NeuroRVQ} Benchmark}
\label{app:neurorvq}

The NeuroRVQ benchmark~\citep{neurorvq} complements our standard benchmark suite with four tasks that the latter does not cover, i.e., four-class executed movement on High-Gamma, target detection on the ERP paradigm of OpenBMI, working-memory load on the digit span recordings of Pavlov et al.~\citep{pavlov2022pupillometry}, and the discrimination of eyes-open from eyes-closed rest on PhysioNet-MI~\citep{schalk2004bci2000}, and it was evaluated under a similar protocol to~\citet{neurorvq}, i.e., 5-fold subject-disjoint cross-validation in which the validation and test folds were distinct, so that each model was trained on about 60\% of the participants. In contrast, \citet{neurorvq} evaluated the same benchmark with 10-fold cross-validation and reported performance on the held-out fold after the last epoch, so that each baseline was trained on 90\% of the participants. We are therefore not directly rank-comparable with the published values, but nevertheless retained a separate validation fold to match the standard split protocol of the remaining benchmarks and to keep the test fold isolated from model selection.

BrainCodes reached the highest mean chance-normalized accuracy on this suite of 0.579, followed by NeuroRVQ at 0.559 and NeuroGPT at 0.484. The proposed model led on executed movement (0.735 against 0.700 for NeuroRVQ), on working memory (0.606 against 0.597 for NeuroGPT) and on resting state, whereas it ranked second on the ERP task. The gap of BrainCodes to older models was larger, e.g., 0.129 in mean chance-normalized accuracy over LaBraM and 0.113 over CBraMod. Notably, we attribute the ERP task lower performance to its short 1-second windows, which yielded only 21 latent frames in BioCodec. As a result, the quantized representation may be able to resolve the evoked response only at a coarse level.

\begin{table}[ht]
\caption{Model performance on the \textsc{NeuroRVQ} benchmark (balanced accuracy). External numbers are as published in~\citet{neurorvq}. Sleep-EDF~\citep{kemp2000analysis} was not included due to computational constraints in both studies. Best per column in \textbf{bold}, second best \underline{underlined}.}
\label{tab:benchmark}
\begin{center}
\begin{footnotesize}
\setlength{\tabcolsep}{4.3pt}
\begin{tabular}{lrcccc>{\columncolor{meanbg}[1.2\tabcolsep][1.25\tabcolsep]}c}
\toprule
\textbf{Methods} & \textbf{Size} & \textbf{Motor} & \textbf{ERP} & \textbf{Memory} & \textbf{Eyes} & \textbf{Mean nBAC} \\
\midrule
BIOT & 3.2M & 0.443 $\pm$ 0.079 & 0.500 $\pm$ 0.000 & 0.510 $\pm$ 0.018 & 0.763 $\pm$ 0.049 & 0.201 $\pm$ 0.037 \\
CBraMod & 4.0M & 0.614 $\pm$ 0.104 & 0.777 $\pm$ 0.052 & 0.574 $\pm$ 0.038 & 0.839 $\pm$ 0.041 & 0.466 $\pm$ 0.052 \\
EEGPT & 25M & 0.313 $\pm$ 0.035 & 0.668 $\pm$ 0.146 & 0.520 $\pm$ 0.017 & 0.797 $\pm$ 0.037 & 0.264 $\pm$ 0.077 \\
NeuroGPT & 79.5M & 0.682 $\pm$ 0.083 & 0.757 $\pm$ 0.048 & \underline{0.597} $\pm$ 0.029 & 0.827 $\pm$ 0.036 & 0.484 $\pm$ 0.043 \\
BrainOmni$^\dagger$ & 33M & 0.585 $\pm$ 0.058 & 0.723 $\pm$ 0.041 & 0.518 $\pm$ 0.023 & 0.852 $\pm$ 0.022 & 0.408 $\pm$ 0.032 \\
LaBraM$^\dagger$ & 5.8M & 0.630 $\pm$ 0.076 & 0.822 $\pm$ 0.040 & 0.526 $\pm$ 0.026 & 0.799 $\pm$ 0.047 & 0.450 $\pm$ 0.042 \\
\textsc{NeuroRVQ}$^\dagger$ & 5.9M & \underline{0.700} $\pm$ 0.073 & \textbf{0.876} $\pm$ 0.033 & 0.574 $\pm$ 0.027 & \underline{0.869} $\pm$ 0.026 & \underline{0.559} $\pm$ 0.035 \\
\midrule
BrainCodes & \textbf{1.9M} & \textbf{0.735} $\pm$ 0.079 & \underline{0.857} $\pm$ 0.032 & \textbf{0.606} $\pm$ 0.022 & \textbf{0.872} $\pm$ 0.025 & \textbf{0.579} $\pm$ 0.035 \\
\bottomrule
\end{tabular}
\end{footnotesize}
\end{center}
\end{table}

\section{Model Performance: Sparse Montage Ablation}

\begin{figure}[ht]
    \centering
    \includegraphics[width=\linewidth]{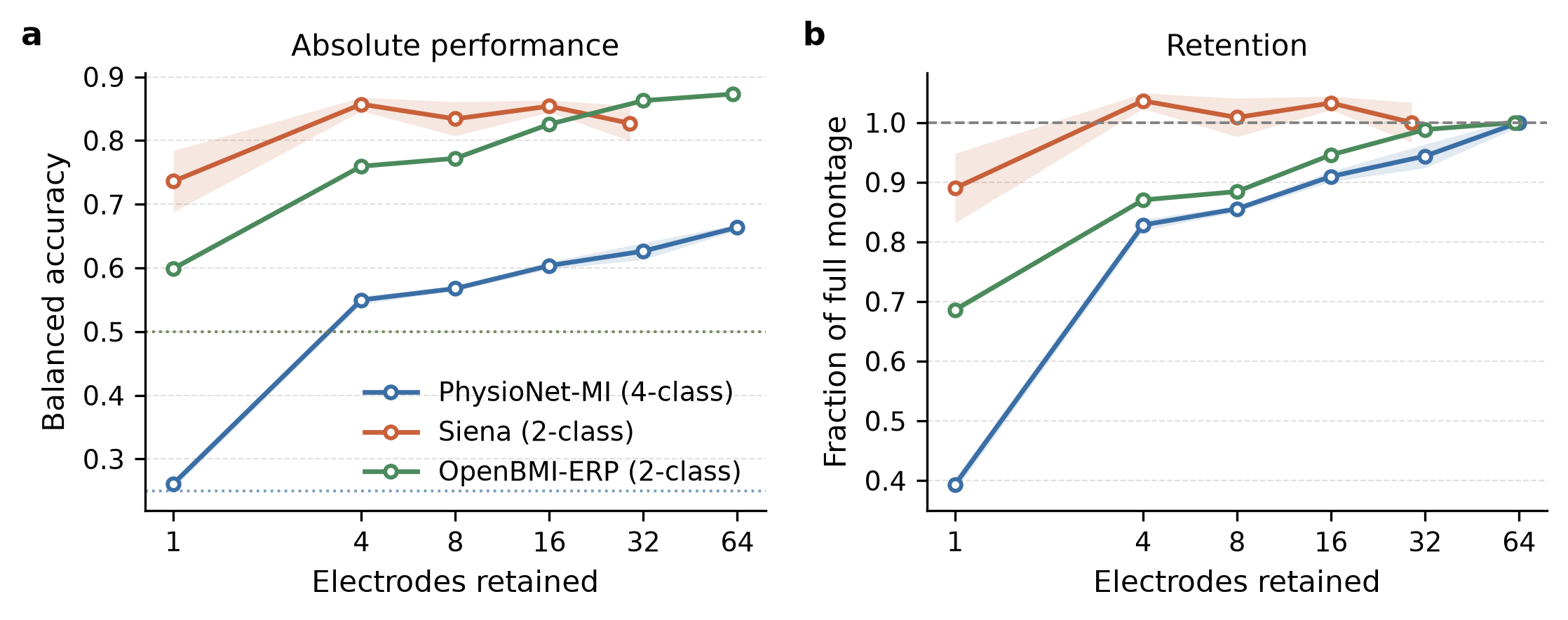}
    \vspace{-8mm}
    \caption{Balanced accuracy as electrodes are removed by farthest-point sampling, with shading marking the standard deviation over seeds. Left: absolute performance, with dotted lines at chance. Right: fraction of each full-montage score, where each full montage sits at its own electrode count.$^\ast$}
    \label{fig:sparse}
\end{figure}

A foundation model that accepts variable electrode montages is most useful if its performance remains stable as spatial sampling becomes increasingly sparse. We therefore evaluated BrainCodes under a controlled montage-reduction protocol designed to separate the effect of electrode count from that of electrode placement. For each of the considered datasets, we constructed nested, task-agnostic subsets of 32, 16, 8, 4 and 1 electrodes using farthest-point sampling over the 3D electrode coordinates. This procedure preserves broad and approximately uniform scalp coverage at every budget while ensuring that each smaller montage is contained within the preceding one. The model was then fine-tuned separately on each reduced montage over three seeds, following the same protocol used for the corresponding full-montage benchmark.

BrainCodes retained 83\% of its full-montage BAC with four electrodes on PhysioNet-MI, 87.0\% on OpenBMI-ERP and scored even better than full fine-tuning on Siena, yielding an average retention of 91\% (Figure~\ref{fig:sparse}b). Performance generally declined gradually as the montage was reduced, with a pronounced drop emerging only in the single-electrode condition. On PhysioNet-MI, for example, balanced accuracy decreased from 0.550 with four electrodes to 0.261 with one, as a single channel cannot preserve the lateralized spatial contrast needed to distinguish imagery of the left and right hands, making this task particularly sensitive to the final reduction. Siena, by contrast, matched its full-montage performance even with four electrodes and none of its reduced-montage results differed significantly from the full montage ($p>0.2$, two-sample $t$-tests across seeds).

\section{Model Performance: Low-resource Fine-tuning}
\label{app:budget}

Label efficiency is commonly summarized by the total number of samples. In EEG, however, the same budget can be allocated either broadly across participants or densely within fewer participants, and these strategies need not be equivalent when label-relevant variability is distributed differently within and across individuals. To separate these effects, we varied participant coverage and within-participant label density independently on PhysioNet-MI and OpenBMI-ERP. PhysioNet-MI crossed participant counts $\{16,24,32,48,60\}$ with $\{24,30,45,60,90\}$ windows per participant, while OpenBMI-ERP crossed $\{8,16,24,32,40\}$ participants with $\{24,30,45,60,90,180,360,720\}$ windows. For each condition, participants were sampled without replacement and windows were sampled within participant without class stratification. BrainCodes was fully fine-tuned over three seeds under our standard protocol, with validation and test partitions held fixed.

\begin{figure}[ht]
    \centering
    \includegraphics[width=
    \linewidth]{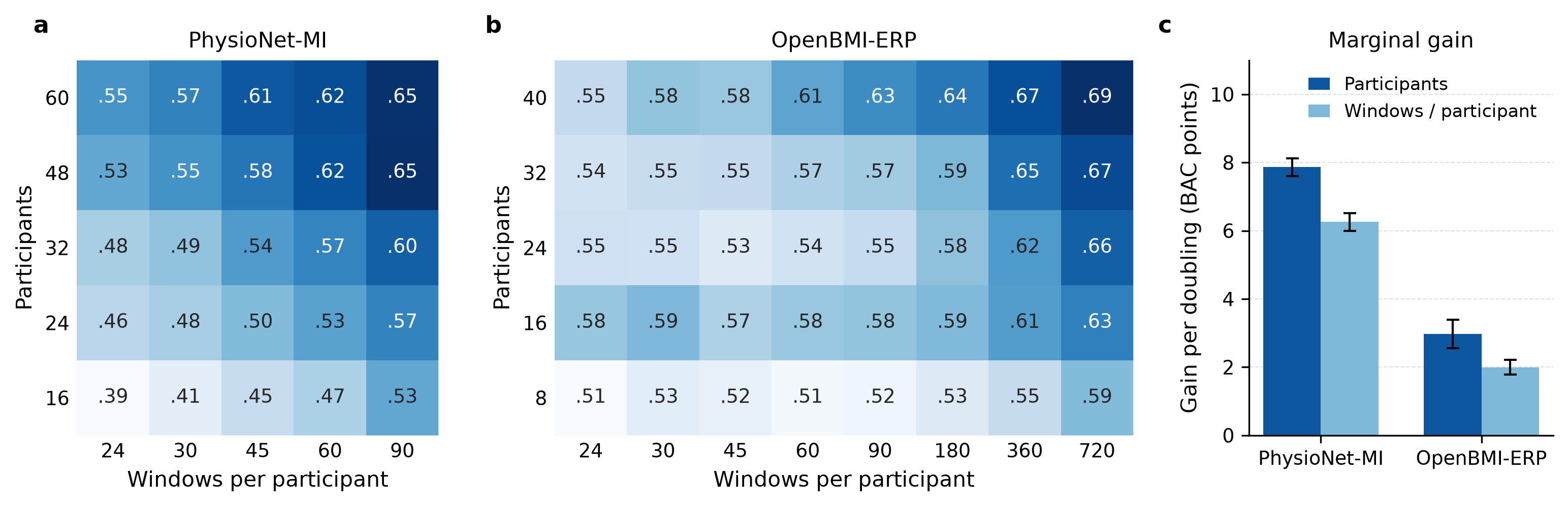}
    \vspace{-3mm}
    \caption{Low-resource fine-tuning under independently varied participant and within-participant label budgets. \textbf{(a,b)} Mean balanced accuracy of BrainCodes across three seeds on PhysioNet-MI and OpenBMI-ERP. \textbf{(c)} Estimated gain in balanced-accuracy points per doubling of each budget axis. Error bars denote coefficient standard errors.}
    \label{fig:label_budget}
\end{figure}

Performance improved along both budget axes, with different sensitivities across datasets (Figure~\ref{fig:label_budget}). On PhysioNet-MI, balanced accuracy rose from 0.386 with 16 participants and 24 windows each to 0.652 with 48 participants and 90 windows each, approaching the full-partition score of 0.664. The progression was smooth across the grid, indicating that both additional participants and additional trials from the same participants contributed consistently. On OpenBMI-ERP, performance increased from 0.511 with 8 participants and 24 windows each to 0.690 with 40 participants and 720 windows each. Gains were modest at the smallest within-participant budgets but became clearer as the number of windows increased. To compare the two budget dimensions directly, we regressed the mean accuracy of each condition on the logarithms of participant count and windows per participant. On PhysioNet-MI, each doubling contributed 7.9 balanced-accuracy points for participants and 6.3 points for windows per participant. The corresponding gains on OpenBMI-ERP were 3.0 and 2.0 points. Broader participant coverage therefore provided the larger marginal return on both datasets, consistent with~\citet{bomatter2026limited}, although dense within-participant annotation remained beneficial and accounted for a substantial fraction of the improvement.

\section{Ablation on the Pretraining Components}

To isolate the contribution of each design choice, we trained 4 otherwise matched variants of BrainCodes: rotary channel encoding was replaced by an MLP over electrode coordinates, block masking by independent random-token masking, BioCodec initialization of the code embeddings by random initialization, and the oscillation-injection objective by no auxiliary objective. Each variant underwent the same pretraining procedure and was then fully fine-tuned on all three datasets. Removing $\mathcal{L}_{\mathrm{osc}}$ caused the largest reduction in mean balanced accuracy, from 0.645 to 0.574, followed by random initialization of the code embeddings at 0.592 (Table~\ref{tab:ablation}). Replacing the positional encoding or masking scheme had smaller effects, yielding 0.613 and 0.623, respectively. Random-token masking was also less stable on BCI IV-2a and EEGMAT. EESM23 was least affected by the positional-encoding change, plausibly because its four in-ear electrodes provide limited spatial structure for either encoding to exploit. Overall, the ablation results in 3 different datasets and setups, show that our pretraining choices provided incremental value to the latent representations of BrainCodes.

\begin{table}[ht]
\caption{Ablation study on the pretraining configuration. We report BAC on three sampled datasets from the main benchmark, with the average computed
across all datasets in three seeds.}
\label{tab:ablation}
\begin{center}
\begin{tabular}{lccc>{\columncolor{meanbg}}c}
\toprule
\textbf{Ablated component} & \textbf{BCI IV-2a} & \textbf{EEGMAT} & \textbf{EESM23} & \textbf{Mean nBAC} \\
\midrule
MLP positional encoding        & 0.631 $\pm$ 0.007          & 0.698 $\pm$ 0.038          & \underline{0.510} $\pm$ 0.010 & 0.439 $\pm$ 0.026 \\
Random token masking           & \textbf{0.640} $\pm$ 0.032 & \underline{0.753} $\pm$ 0.066 & 0.477 $\pm$ 0.038          & \underline{0.466} $\pm$ 0.049 \\
Codebook initialization        & 0.639 $\pm$ 0.020          & 0.677 $\pm$ 0.034          & 0.461 $\pm$ 0.026          & 0.409 $\pm$ 0.026 \\
Auxiliary objective $\mathcal{L}_{\mathrm{osc}}$ & 0.607 $\pm$ 0.025 & 0.670 $\pm$ 0.032 & 0.444 $\pm$ 0.021 & 0.383 $\pm$ 0.025 \\
\midrule
Official BrainCodes            & \underline{0.639} $\pm$ 0.002 & \textbf{0.772} $\pm$ 0.015 & \textbf{0.525} $\pm$ 0.025 & \textbf{0.498} $\pm$ 0.014 \\
\bottomrule
\end{tabular}
\end{center}
\end{table}

We also separated the contributions of the pretrained tokenizer and backbone. Retaining BioCodec but initializing the backbone from scratch reduced mean chance-normalized BAC from 0.498 to 0.270 (Table~\ref{tab:withheld}). Replacing BioCodec with a randomly initialized quantizer reduced the mean further to 0.121, with EEGMAT falling to chance. Under the randomly initialized backbone, BioCodec improved balanced accuracy by 0.212 on BCI IV-2a, 0.060 on EEGMAT and 0.037 on EESM23. Thus, backbone pretraining accounted for the largest share of the gain, while the BioCodec tokenizer provided an additional advantage that we found to be strongest for motor imagery. Together with Table~\ref{tab:ablation}, these results identify the oscillation objective and tokenizer initialization as the most consequential individual components of our pretraining method.

\begin{table}[ht]
\caption{Contributions of tokenizer and backbone pretraining. Values are fully fine-tuned mean balanced accuracy $\pm$ standard deviation over 3 seeds. The last column reports mean chance-normalized balanced accuracy. \textit{Random} indicates an identical to BioCodec but randomly initialized RVQ.}
\label{tab:withheld}
\begin{center}
\begin{tabular}{llccc>{\columncolor{meanbg}}c}
\toprule
\textbf{Tokenizer} & \textbf{Backbone} & \textbf{BCI IV-2a} & \textbf{EEGMAT} & \textbf{EESM23} & \textbf{Mean nBAC} \\
\midrule
BioCodec & BrainCodes   & \textbf{0.639} $\pm$ 0.002 & \textbf{0.772} $\pm$ 0.015 & \textbf{0.525} $\pm$ 0.025 & \textbf{0.498} $\pm$ 0.014 \\
BioCodec & from scratch & 0.531 $\pm$ 0.015 & 0.561 $\pm$ 0.047 & 0.429 $\pm$ 0.019 & 0.270 $\pm$ 0.033 \\
Random    & from scratch & 0.319 $\pm$ 0.010 & 0.501 $\pm$ 0.002 & 0.392 $\pm$ 0.030 & 0.121 $\pm$ 0.013 \\
\bottomrule
\end{tabular}
\end{center}
\end{table}

\section{The Learned Channel Positional Encoding}

BrainCodes learns how differences in three-dimensional electrode position should affect channel attention. For attention head $h$ and rotary pair $r$, a learned vector $\boldsymbol{\omega}_{h,r}\in\mathbb{R}^{3}$ converts the displacement between electrodes $i$ and $j$ into a relative phase,
\begin{equation}
    \Delta\phi_{h,r}(i,j)
    =
    \boldsymbol{\omega}_{h,r}^{\top}
    (\mathbf{p}_i-\mathbf{p}_j).
\end{equation}
The direction of $\boldsymbol{\omega}_{h,r}$ determines the spatial axis to which the pair responds, while its magnitude determines how rapidly the phase changes with distance. A near-zero vector is therefore insensitive to electrode position, whereas a large vector produces stronger spatial differentiation. BrainCodes learns 8 such pairs for each of its eight attention heads, initialized at geometrically decreasing magnitudes. The same learned rotations are shared across all channel-attention layers, so each head applied a consistent spatial code throughout the backbone.

\begin{figure}[ht]
    \centering
    \includegraphics[width=0.84\linewidth]{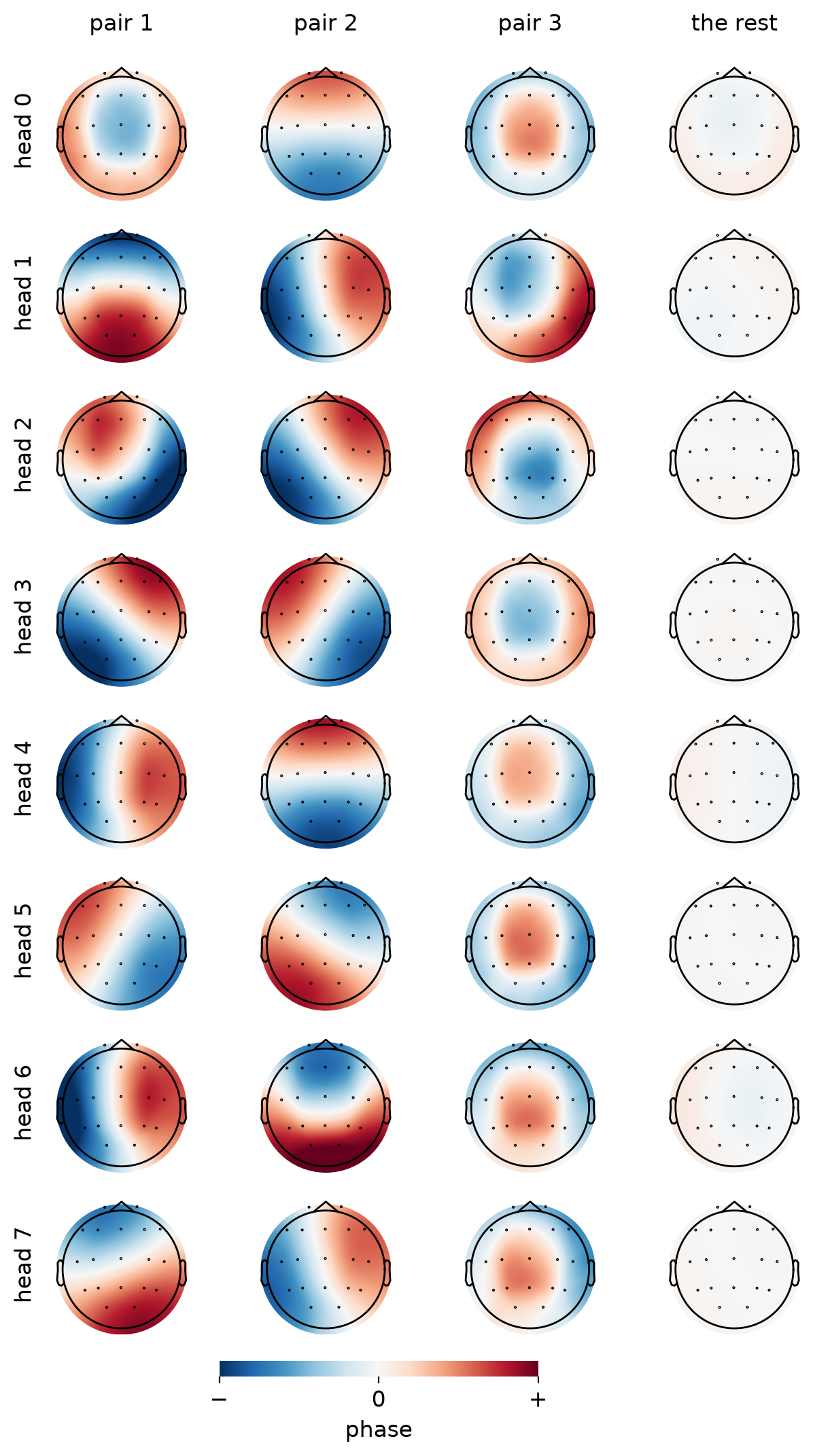}
    \vspace{-4mm}
    \caption{Phase assigned by the learned channel rotary encoding to a 10--20 montage, for each of the eight attention heads. Columns: the three rotation pairs of largest frequency magnitude, ordered by decreasing magnitude, and the mean over the remaining five pairs.}
    \label{fig:rope_per_head}
\end{figure}

Pretraining retained a compact set of spatial modes. Of the 64 head--pair vectors, 26 grew beyond their initial magnitudes, reaching norms of 0.907--1.878, increased by a median of 2.85 times. Most of this spatial structure was concentrated in three rotary pairs per head; the remaining pairs produced little positional variation (Figure~\ref{fig:rope_per_head}). The resulting positional similarity remained between 0.61 and 0.70 even at the largest possible inter-electrode distance, indicating a smooth spatial bias rather than a hard separation between distant channels. Within each attention head, the retained directions formed an approximately orthogonal three-dimensional frame: the mean ratio between their smallest and largest singular values was 0.770, compared with 0.257 for size-matched random sets ($p<0.001$). Collectively, these directions covered the left--right, anterior--posterior and inferior--superior axes evenly. Their normalized spatial gains were also nearly isotropic (0.353, 0.360 and 0.286), despite the unequal geometric extent of the montage along those axes (0.491, 0.358 and 0.150).

\begin{figure}[ht]
    \centering
    \includegraphics[width=0.85\linewidth]{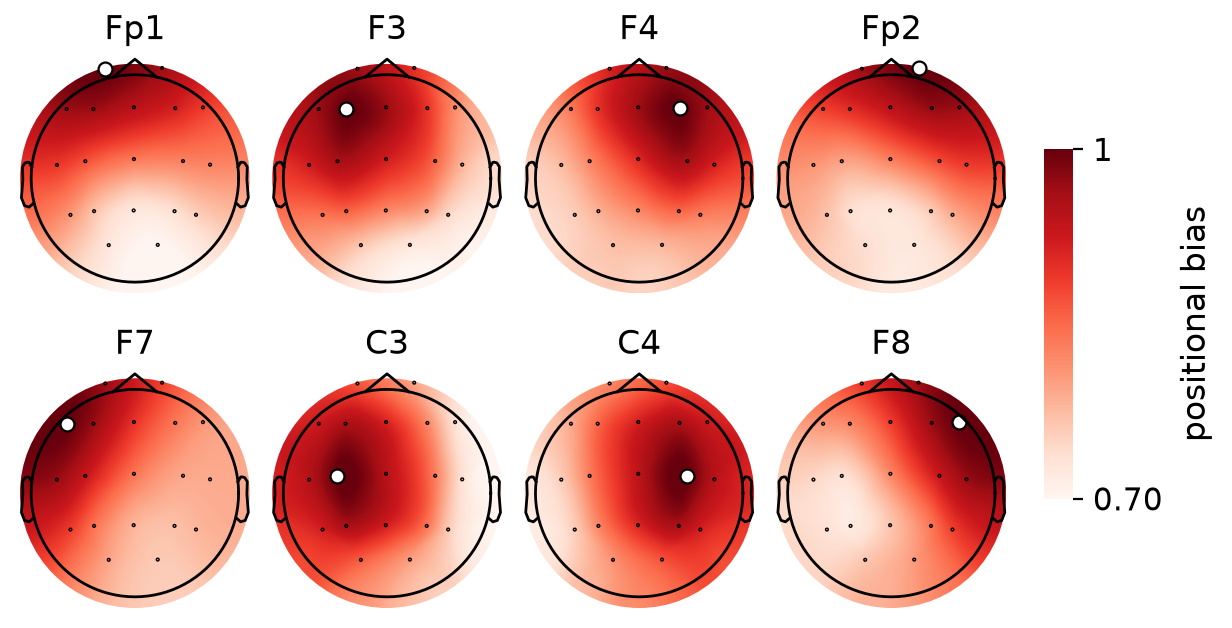}
    \vspace{-2mm}
    \caption{Positional bias of the learned channel rotary encoding for a sample attention head from BrainCodes, seeded at each of eight electrodes, where the white marker gives the seed and the color gives the bias assigned to every other site. \textbf{Top}: frontal. \textbf{Bottom}: fronto-temporal and central.}
    \label{fig:rope_seed_grid}
\end{figure}

Overall, pretraining transformed the initial multiscale encoding into a compact, approximately isotropic three-dimensional distance code that was replicated across attention heads. The resulting bias depends primarily on relative electrode separation rather than absolute scalp location, so seeding it at different electrodes produces similar smoothly decaying spatial fields (Figure~\ref{fig:rope_seed_grid}). This geometry is consistent with the isotropic spatial mixing produced by volume conduction.

\begin{figure}[t]
    \centering
    \includegraphics[width=0.9\linewidth]{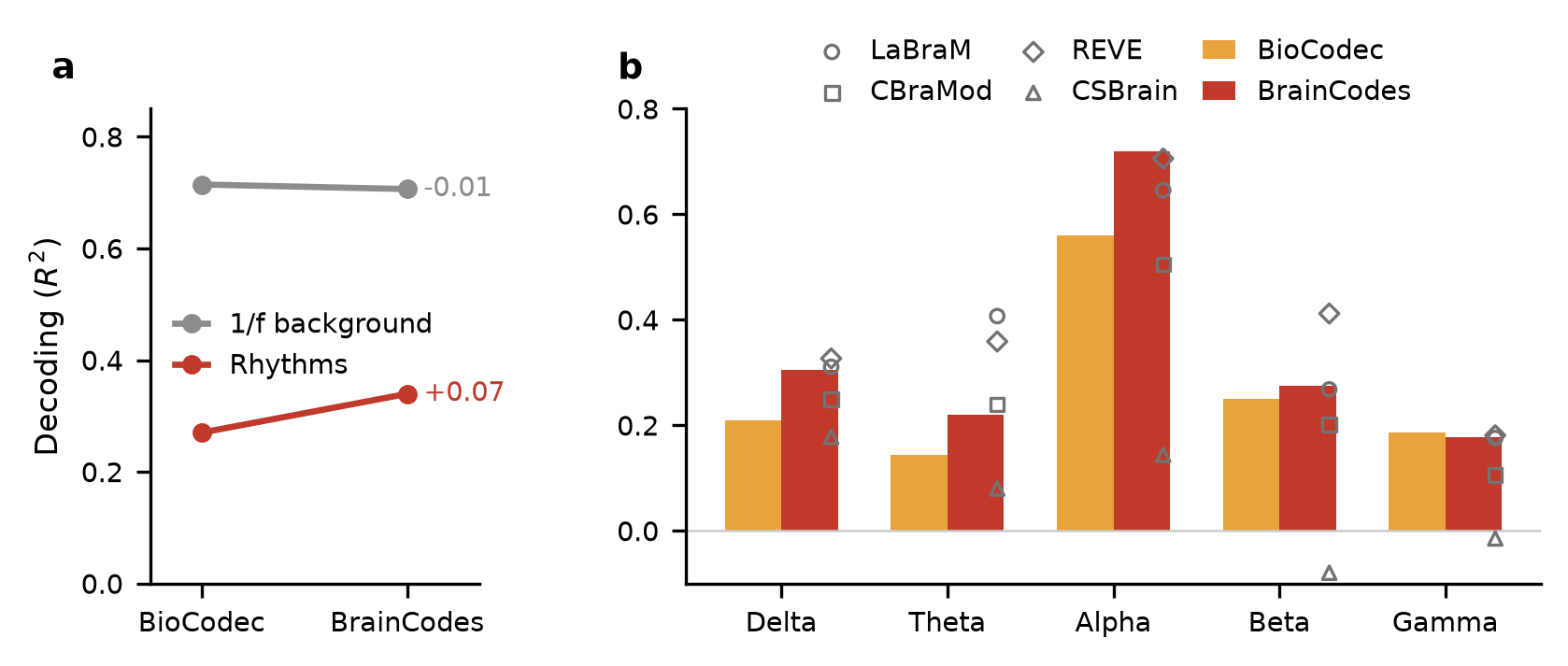}
    \vspace{-2mm}
    \caption{Linear decoding of aperiodic and oscillatory spectral components from frozen representations on PhysioNet-MI, measured by within-channel $R^2$. \textbf{(a)} Decoding of the $1/f$ exponent and mean oscillatory power across the five canonical frequency bands from BioCodec and BrainCodes. \textbf{(b)} Band-specific decoding, with other EEG foundation models shown as reference markers.}    \label{fig:natural_rhythms}
\end{figure}

\section{Spectral Sensitivity Analysis}
\label{app:spectral}

Since the auxiliary objective explicitly targeted narrowband activity, we tested whether BrainCodes makes naturally occurring EEG rhythms more accessible without simply encoding the broadband spectral slope. We analyzed 800 PhysioNet-MI windows from 19 participants and evaluated on 800 windows from 20 separate participants. For every channel, the log spectrum from 1 to 45\,Hz was decomposed into periodic and aperiodic components following \citet{donoghue2020parameterizing}. Specifically, we fitted a line in log--log coordinates and iteratively refitted it using bins below the 40th percentile of the residual distribution, preventing spectral peaks from biasing the aperiodic fit. Its slope defined the $1/f$ exponent, while the mean residual log power within each canonical frequency band defined the oscillatory targets. Linear regression heads were fitted to the time-averaged tokens of each frozen model. Targets were centered within each channel using only the fitting set, hence preventing electrode identity from supporting prediction.

Relative to its BioCodec input, the BrainCodes bottleneck increased mean oscillatory decoding from $R^2=0.270$ to 0.340, while decoding of the $1/f$ exponent remained effectively unchanged (Figure~\ref{fig:natural_rhythms}a). The improvement was concentrated in the lower-frequency rhythms: $R^2$ increased by 0.096 in delta, 0.075 in theta and 0.160 in alpha, significantly above zero by participant-bootstrap confidence intervals. Differences in beta, gamma and the aperiodic exponent were not significant. The delta improvement was notable because this band fell outside the 4--32\,Hz range used for burst injection during pretraining, suggesting that the learned sensitivity generalized beyond the frequencies explicitly supervised. Across all six models, BrainCodes yielded the strongest alpha decoding ($R^2=0.720$), compared with 0.706 for REVE and 0.645 for LaBraM (Figure~\ref{fig:natural_rhythms}b). REVE and LaBraM nevertheless decoded oscillatory power more strongly on average (0.397 and 0.361), and REVE best captured the $1/f$ exponent (0.756). Thus, BrainCodes selectively made low-frequency oscillatory structure more linearly accessible than its tokenizer did, while leaving the already strong representation of the aperiodic component essentially unchanged. Importantly, this pattern was observed in real EEG rather than synthetically perturbed proxies.

\section{Signal Capacity \& Model Scaling}

BrainCodes is designed around two forms of efficiency: a compressed discrete (code) representation and a lightweight U-Net backbone. The BioCodec tokenizer imposes an explicit information bottleneck, encoding approximately \(1{,}000\) bits/s per channel compared with \(4{,}000\) bits/s for the original 16-bit signal. However, we showed in Table~\ref{tab:withheld} that the tokenizer outperformed a random quantizer at the same bitrate and, after pretraining, BrainCodes surpassed the random projection on all considered datasets. Thus, the discrete bottleneck reduces input capacity without constraining the performance of the pretrained model.

We did not directly study parameter scaling and BrainCodes was pretrained at a single capacity of 1.9M parameters. Existing results nevertheless support this design choice. Scaling LaBraM from 5.8M to 369M parameters improved TUEV balanced accuracy from 0.6409 to 0.6616~\citep{labram}, which corresponds to less than a participant-level bootstrap standard deviation. Notably, participant-level resampling produced 9.2 times greater variability than window-level resampling, indicating that the apparent scaling gain is small relative to uncertainty across individuals. CBraMod likewise found that performance improved with model size, while gains from additional pretraining data slowed beyond 1,000 hours~\citep{cbramod}. On the other hand, REVE reported continued linear-probing gains from 12M to 408M parameters~\citep{reve}. Together with our finding that increasing participant coverage yields greater returns than collecting more data per participant (Appendix~\ref{app:budget}), these results motivated a complementary focus on efficiency.

\section{Model Performance: Detailed Results}

The tables of this section report balanced accuracy together with Cohen's $\kappa$ and weighted F1 for multi-class tasks, or with AUROC and AUPRC for binary tasks, as mean and standard deviation over three seeds. Baseline entries were quoted from the publications of CBraMod~\citep{cbramod}, REVE~\citep{reve} and LaBraM~\citep{labram} where a directly comparable protocol existed, and the EESM23 baselines of Table~\ref{tab:eesm} were quoted from TFM-Tokenizer~\citep{tfmtokenizer}, whose participant split we adopted. All datasets were resampled to 250\,Hz.

\textbf{TUAB}~\citep{obeid2016temple}. The Temple University Hospital Abnormal EEG Corpus contains routine clinical recordings from 2,329 participants, each labeled by clinical experts as normal or abnormal at the session level. Recordings were reduced to 21 channels of the 10--20 system, band-passed between 0.5 and 75\,Hz, and segmented into 10-second windows. We followed the provided training-test split
and applied the same validation split as in~\citet{reve}.

\begin{table}[ht]
\caption{Model performance on abnormality detection (TUAB, 2-class).}
\begin{center}
\begin{tabular}{lcccc}
\toprule
\textbf{Methods} & \textbf{Size} & \textbf{Balanced Accuracy} & \textbf{AUC-PR} & \textbf{AUROC} \\
\midrule
EEGNet          & 0.1M    & 0.7642 $\pm$ 0.0036 & 0.8299 $\pm$ 0.0043 & 0.8412 $\pm$ 0.0031 \\
EEGConformer    & 0.8M    & 0.7758 $\pm$ 0.0049 & 0.8427 $\pm$ 0.0054 & 0.8445 $\pm$ 0.0038 \\
SPaRCNet        & 0.8M  & 0.7896 $\pm$ 0.0018 & 0.8414 $\pm$ 0.0018 & 0.8676 $\pm$ 0.0012 \\
ContraWR        & 1.6M  & 0.7746 $\pm$ 0.0041 & 0.8421 $\pm$ 0.0104 & 0.8456 $\pm$ 0.0074 \\
CNN-Transformer & 3.2M    & 0.7777 $\pm$ 0.0022 & 0.8433 $\pm$ 0.0039 & 0.8461 $\pm$ 0.0013 \\
ST-Transformer  & 2.6M    & 0.7966 $\pm$ 0.0023 & 0.8521 $\pm$ 0.0026 & 0.8707 $\pm$ 0.0019 \\
\midrule
BIOT            & 3.2M  & 0.7959 $\pm$ 0.0057 & 0.8792 $\pm$ 0.0023 & 0.8815 $\pm$ 0.0043 \\
LaBraM-Base     & 5.8M  & 0.8140 $\pm$ 0.0019 & 0.8965 $\pm$ 0.0016 & 0.9022 $\pm$ 0.0009 \\
CBraMod         & 4.0M  & 0.8289 $\pm$ 0.0022 & 0.9258 $\pm$ 0.0008 & 0.9227 $\pm$ 0.0011 \\
REVE-Base       & 69M   & \textbf{0.8315} $\pm$ 0.0014 & \textbf{0.9281} $\pm$ 0.0009 & \textbf{0.9245} $\pm$ 0.0013 \\
\midrule
BrainCodes & \textbf{1.9M} & 0.8110 $\pm$ 0.0125 & 0.8820 $\pm$ 0.0121 & 0.8833 $\pm$ 0.0123 \\
\bottomrule
\end{tabular}
\end{center}
\end{table}

\textbf{TUEV}~\citep{obeid2016temple}: The Temple University Hospital - Epilepsy EEG dataset (TUEV) is another popular subset of TUEG. The dataset provides a large-scale EEG resource to advance the understanding of epileptic activity. In our study we used 6 expert-annotated event classes: background, artifact, eye movement, generalized periodic epileptiform discharges, periodic lateralized epileptiform discharges, and spike and sharp wave. For consistency with prior studies, we preprocessed the dataset to 16 bipolar channels in the 10-20 system, band-pass filtered at 0.5–75 Hz, and removed powerline noise with a 60\,Hz notch. The dataset is split into 5-second samples. We followed the provided training-test split and applied the same validation split and positional encoding as in~\citet{reve}, i.e., we used the average 3D position of each bipolar montage.

\begin{table}[t]
\caption{Model performance on event type classification (TUEV, 6-class).}
\begin{center}
\begin{tabular}{lrccc}
\toprule
\textbf{Methods} & \textbf{Size} & \textbf{Balanced Accuracy} & \textbf{Cohen's $\kappa$} & \textbf{Weighted F1} \\
\midrule
EEGNet          & 0.1M    & 0.3876 $\pm$ 0.0143 & 0.3577 $\pm$ 0.0155 & 0.6539 $\pm$ 0.0120 \\
EEGConformer    & 0.8M    & 0.4074 $\pm$ 0.0164 & 0.3967 $\pm$ 0.0195 & 0.6983 $\pm$ 0.0152 \\
SPaRCNet        & 0.8M  & 0.4161 $\pm$ 0.0262 & 0.4233 $\pm$ 0.0181 & 0.7024 $\pm$ 0.0104 \\
ContraWR        & 1.6M  & 0.4384 $\pm$ 0.0349 & 0.3912 $\pm$ 0.0237 & 0.6893 $\pm$ 0.0136 \\
CNN-Transformer & 3.2M    & 0.4087 $\pm$ 0.0161 & 0.3815 $\pm$ 0.0134 & 0.6854 $\pm$ 0.0293 \\
ST-Transformer  & 2.6M    & 0.3984 $\pm$ 0.0228 & 0.3765 $\pm$ 0.0306 & 0.6823 $\pm$ 0.0190 \\
\midrule
BIOT            & 3.2M  & 0.5281 $\pm$ 0.0225 & 0.5273 $\pm$ 0.0249 & 0.7492 $\pm$ 0.0082 \\
LaBraM-Base     & 5.8M  & 0.6409 $\pm$ 0.0065 & 0.6637 $\pm$ 0.0093 & 0.8312 $\pm$ 0.0052 \\
LaBraM-Large    & 46M   & 0.6581 $\pm$ 0.0156 & 0.6622 $\pm$ 0.0136 & 0.8315 $\pm$ 0.0040 \\
CBraMod         & 4.0M  & 0.6671 $\pm$ 0.0107 & 0.6772 $\pm$ 0.0096 & 0.8342 $\pm$ 0.0064 \\
REVE-Base       & 69M   & 0.6759 $\pm$ 0.0229 & 0.6783 $\pm$ 0.0199 & 0.8451 $\pm$ 0.0129 \\
\midrule
BrainCodes & \textbf{1.9M} & \textbf{0.6903} $\pm$ 0.0121 & \textbf{0.7093} $\pm$ 0.0076 & \textbf{0.8477} $\pm$ 0.0049 \\
\bottomrule
\end{tabular}
\end{center}
\end{table}

\textbf{ISRUC}~\citep{khalighi2016isruc} contains overnight polysomnography from adults with sleep disorders. Each 30-second epoch was scored by experts into wake, N1, N2, N3 and REM. Similarly to the compared models, we use Subgroup 1, which contains recordings from 100 subjects. EEG signals are resampled to 250\,Hz, segmented into 30-second epochs, each labeled with one of the 5 sleep stages. Subjects 1–80 were used for training, 81–90 for validation, and 91–100
for testing.

\begin{table}[ht]
\caption{Model performance on sleep staging (ISRUC, 5-class).}
\begin{center}
\begin{tabular}{lrccc}
\toprule
\textbf{Methods} & \textbf{Size} & \textbf{Balanced Accuracy} & \textbf{Cohen's Kappa} & \textbf{Weighted F1} \\
\midrule
EEGNet          & 0.1M    & 0.7154 $\pm$ 0.0121 & 0.7040 $\pm$ 0.0173 & 0.7513 $\pm$ 0.0124 \\
EEGConformer    & 0.8M    & 0.7400 $\pm$ 0.0133 & 0.7143 $\pm$ 0.0162 & 0.7634 $\pm$ 0.0151 \\
SPaRCNet        & 0.8M  & 0.7487 $\pm$ 0.0075 & 0.7097 $\pm$ 0.0132 & 0.7624 $\pm$ 0.0092 \\
ContraWR        & 1.6M  & 0.7402 $\pm$ 0.0126 & 0.7178 $\pm$ 0.0156 & 0.7610 $\pm$ 0.0137 \\
CNN-Transformer & 3.2M    & 0.7363 $\pm$ 0.0087 & 0.7129 $\pm$ 0.0121 & 0.7719 $\pm$ 0.0105 \\
ST-Transformer  & 2.6M    & 0.7381 $\pm$ 0.0205 & 0.7013 $\pm$ 0.0352 & 0.7681 $\pm$ 0.0175 \\
DeepSleepNet    & 25M    & 0.7419 $\pm$ 0.0144 & 0.7036 $\pm$ 0.0241 & 0.7643 $\pm$ 0.0122 \\
USleep          & 3.1M    & 0.7586 $\pm$ 0.0116 & 0.7209 $\pm$ 0.0143 & 0.7805 $\pm$ 0.0105 \\
\midrule
BIOT            & 3.2M  & 0.7527 $\pm$ 0.0121 & 0.7192 $\pm$ 0.0231 & 0.7790 $\pm$ 0.0146 \\
LaBraM-Base     & 5.8M  & 0.7633 $\pm$ 0.0102 & 0.7231 $\pm$ 0.0182 & 0.7810 $\pm$ 0.0133 \\
CBraMod         & 4.0M  & 0.7865 $\pm$ 0.0110 & 0.7442 $\pm$ 0.0152 & 0.8011 $\pm$ 0.0099 \\
REVE-Base       & 69M   & 0.7819 $\pm$ 0.0078 & 0.7500 $\pm$ 0.0156 & 0.8005 $\pm$ 0.0130 \\
\midrule
BrainCodes & \textbf{1.9M} & \textbf{0.7960} $\pm$ 0.0078 & \textbf{0.7513} $\pm$ 0.0265 & \textbf{0.8097} $\pm$ 0.0225 \\
\bottomrule
\end{tabular}
\end{center}
\end{table}

\textbf{Siena}~\citep{detti2020eeg}: The Siena Scalp EEG Database contains EEG recordings from 14 patients with epilepsy, with electrodes placed according to the 10--20 system and sampled at 512\,Hz, with the onset and offset of every seizure annotated by clinical experts. We retained the 29 EEG channels that were consistently available across all patients, resampled to 250\,Hz and segmented it into 10-second windows labeled as \textit{ictal} or \textit{non-ictal}, with no patient contributing to more than one partition. We employed a subject-disjoint split: all data from the last 2 subjects (PN16 and PN17) were the test set, while the remaining 12 patients were split into train (first 9) and validation (last 3) sets.

\begin{table}[ht]
\caption{Model performance on seizure detection (Siena, 2-class).}
\begin{center}
\begin{tabular}{lrccc}
\toprule
\textbf{Methods} & \textbf{Size} & \textbf{Balanced Accuracy} & \textbf{AUC-PR} & \textbf{AUROC} \\
\midrule
EEGNet          & 0.1M  & 0.7487 $\pm$ 0.0521 & 0.3753 $\pm$ 0.0867 & 0.8687 $\pm$ 0.0527 \\
EEGConformer    & 0.8M  & 0.7556 $\pm$ 0.0210 & 0.2091 $\pm$ 0.0786 & 0.8159 $\pm$ 0.0261 \\
SPaRCNet        & 0.8M  & 0.6572 $\pm$ 0.0381 & 0.3164 $\pm$ 0.0659 & 0.7334 $\pm$ 0.0857 \\
ContraWR        & 1.6M  & 0.6546 $\pm$ 0.0311 & 0.3711 $\pm$ 0.0405 & 0.7819 $\pm$ 0.0596 \\
CNN-Transformer & 3.2M  & 0.6982 $\pm$ 0.0560 & 0.3835 $\pm$ 0.0709 & 0.8719 $\pm$ 0.0403 \\
ST-Transformer  & 2.6M  & 0.7527 $\pm$ 0.0381 & 0.3636 $\pm$ 0.0252 & 0.8884 $\pm$ 0.0091 \\
\midrule
BIOT            & 3.2M  & 0.7352 $\pm$ 0.0669 & 0.3809 $\pm$ 0.0892 & 0.9029 $\pm$ 0.0304 \\
LaBraM-Base     & 5.8M  & 0.7082 $\pm$ 0.0329 & 0.3122 $\pm$ 0.0976 & 0.8814 $\pm$ 0.0328 \\
CBraMod         & 4.0M  & 0.7317 $\pm$ 0.0647 & 0.4107 $\pm$ 0.0720 & 0.9038 $\pm$ 0.0218 \\
REVE-Base       & 69M   & 0.7400 $\pm$ 0.0070   & 0.3810 $\pm$ 0.0530   & 0.8750 $\pm$ 0.0300 \\ 
CSBrain         & 8.9M  & 0.7662 $\pm$ 0.0471 & 0.4871 $\pm$ 0.0343 & \textbf{0.9076} $\pm$ 0.0119 \\
\midrule
BrainCodes      & 1.9M  & \textbf{0.8270} $\pm$ 0.0340 & \textbf{0.5427} $\pm$ 0.1030 & 0.8797 $\pm$ 0.0232 \\
\bottomrule
\end{tabular}
\end{center}
\end{table}

\paragraph{PhysioNet-MI}~\citep{schalk2004bci2000, pollard_physionet_2026}: The PhysioNet Motor Imagery database represents a foundation resource for studying brain–computer interface using EEG. The dataset consists of over 1,500 EEG recordings from 109 healthy subjects acquired through the BCI2000 system.
Each recording was obtained with 64 electrodes at a sampling rate of 160 Hz. The participants were instructed to complete 14 experimental runs, including motor execution and motor imagery. In our study we incorporate the 4-class setup on imagery as well as the detection of \textit{eyes-open} vs. \textit{eyes-closed} state, as benchmarked by~\citet{neurorvq}.

\begin{table}[ht]
\caption{Model performance on motor imagery (PhysioNet-MI, 4-class).}
\begin{center}
\begin{tabular}{lrccc}
\toprule
\textbf{Methods} & \textbf{Size} & \textbf{Balanced Accuracy} & \textbf{Cohen's $\kappa$} & \textbf{Weighted F1} \\
\midrule
EEGNet          & 0.1M    & 0.5814 $\pm$ 0.0125 & 0.4468 $\pm$ 0.0199 & 0.5796 $\pm$ 0.0115 \\
EEGConformer    & 0.8M    & 0.6049 $\pm$ 0.0104 & 0.4736 $\pm$ 0.0171 & 0.6062 $\pm$ 0.0095 \\
SPaRCNet        & 0.8M & 0.5932 $\pm$ 0.0152 & 0.4564 $\pm$ 0.0234 & 0.5937 $\pm$ 0.0147 \\
ContraWR        & 1.6M  & 0.5892 $\pm$ 0.0133 & 0.4527 $\pm$ 0.0248 & 0.5918 $\pm$ 0.0116 \\
CNN-Transformer & 3.2M    & 0.6053 $\pm$ 0.0118 & 0.4725 $\pm$ 0.0223 & 0.6041 $\pm$ 0.0105 \\
ST-Transformer  & 2.6M    & 0.6035 $\pm$ 0.0081 & 0.4712 $\pm$ 0.0199 & 0.6053 $\pm$ 0.0075 \\
\midrule
BIOT            & 3.2M  & 0.6153 $\pm$ 0.0154 & 0.4875 $\pm$ 0.0272 & 0.6158 $\pm$ 0.0197 \\
LaBraM-Base     & 5.8M  & 0.6173 $\pm$ 0.0122 & 0.4912 $\pm$ 0.0192 & 0.6177 $\pm$ 0.0141 \\
CBraMod         & 4.0M  & 0.6417 $\pm$ 0.0091 & 0.5222 $\pm$ 0.0169 & 0.6427 $\pm$ 0.0100 \\
REVE-Base       & 69M    & 0.6480 $\pm$ 0.0140 & 0.5306 $\pm$ 0.0187 & 0.6484 $\pm$ 0.0170 \\
\midrule
BrainCodes      & \textbf{1.9M}  & \textbf{0.6637} $\pm$ 0.0059 & \textbf{0.5510} $\pm$ 0.0079 & \textbf{0.6640} $\pm$ 0.0044 \\
\bottomrule
\end{tabular}
\end{center}
\end{table}

\paragraph{BCI IV-2a}~\citep{brunner2008bci}: The BCI Competition 2008 – Graz dataset A (BCI IV-2a) includes multi-session EEG recordings from nine subjects engaged in cue-based motor imagery experiments. Specifically, each subject was instructed to perform four imagined movements: left hand, right hand, feet, and tongue. Each participant visited the research lab on distinct days to record separate sessions, where each session included six runs and 48 trials per run (12 trials per movement). The EEG data was acquired through a 22-electrode EEG equipment at 250 Hz.

\begin{table}[ht]
\caption{Model performance on motor imagery (BCIC-IV-2a, 4-class).}
\begin{center}
\begin{tabular}{lrccc}
\toprule
\textbf{Methods} & \textbf{Size} & \textbf{Balanced Accuracy} & \textbf{Cohen's $\kappa$} & \textbf{Weighted F1} \\
\midrule
EEGNet          & 0.1M    & 0.4482 $\pm$ 0.0094 & 0.2693 $\pm$ 0.0121 & 0.4226 $\pm$ 0.0108 \\
EEGConformer    & 0.8M    & 0.4696 $\pm$ 0.0106 & 0.2924 $\pm$ 0.0141 & 0.4533 $\pm$ 0.0128 \\
SPaRCNet        & 0.8M & 0.4635 $\pm$ 0.0117 & 0.2847 $\pm$ 0.0147 & 0.4432 $\pm$ 0.0126 \\
ContraWR        & 1.6M  & 0.4678 $\pm$ 0.0125 & 0.2905 $\pm$ 0.0160 & 0.4413 $\pm$ 0.0142 \\
CNN-Transformer & 3.2M    & 0.4600 $\pm$ 0.0108 & 0.2800 $\pm$ 0.0148 & 0.4460 $\pm$ 0.0114 \\
ST-Transformer  & 2.6M    & 0.4575 $\pm$ 0.0145 & 0.2733 $\pm$ 0.0198 & 0.4471 $\pm$ 0.0142 \\
\midrule
BIOT            & 3.2M  & 0.4748 $\pm$ 0.0093 & 0.2997 $\pm$ 0.0139 & 0.4607 $\pm$ 0.0125 \\
LaBraM-Base     & 5.8M  & 0.4869 $\pm$ 0.0085 & 0.3159 $\pm$ 0.0154 & 0.4758 $\pm$ 0.0103 \\
CBraMod         & 4.0M  & 0.5138 $\pm$ 0.0066 & 0.3518 $\pm$ 0.0094 & 0.4984 $\pm$ 0.0085 \\
REVE-Base       & 69M    & \textbf{0.6396} $\pm$ 0.0095 & \textbf{0.5194} $\pm$ 0.0126 & 0.6339 $\pm$ 0.0110 \\
\midrule
BrainCodes      & \textbf{1.9M}  & 0.6392 $\pm$ 0.0015 & 0.5189 $\pm$ 0.0020 & \textbf{0.6347} $\pm$ 0.0005 \\
\bottomrule
\end{tabular}
\end{center}
\end{table}

\textbf{EEGMAT}~\citep{zyma2019electroencephalograms}: The Mental Arithmetic dataset contains EEG recordings from 36 subjects, labeled as \textit{with} or \textit{without} mental stress depending on whether a mental arithmetic task was being performed. Signals were recorded from 20 electrodes (10–20 system) at 500 Hz, band-pass filtered (0.5–45 Hz), resampled to 250 Hz, and segmented into 5-second windows. Subjects 1–28 are used for training, 29–32 for validation, and 33–36 for testing, as in~\citet{reve}.

\begin{table}[ht]
\caption{Model performance on mental stress detection (EEGMAT, 2-class).}
\begin{center}
\begin{tabular}{lrccc}
\toprule
\textbf{Methods} & \textbf{Size} & \textbf{Balanced Accuracy} & \textbf{AUC-PR} & \textbf{AUROC} \\
\midrule
EEGNet          & 0.1M    & 0.6770 $\pm$ 0.0116 & 0.5763 $\pm$ 0.0102 & 0.7321 $\pm$ 0.0108 \\
EEGConformer    & 0.8M    & 0.6805 $\pm$ 0.0123 & 0.5829 $\pm$ 0.0134 & 0.7424 $\pm$ 0.0128 \\
SPaRCNet        & 0.8M  & 0.6879 $\pm$ 0.0107 & 0.5825 $\pm$ 0.0193 & 0.7418 $\pm$ 0.0132 \\
ContraWR        & 1.6M  & 0.6631 $\pm$ 0.0097 & 0.5787 $\pm$ 0.0164 & 0.7332 $\pm$ 0.0082 \\
CNN-Transformer & 3.2M    & 0.6779 $\pm$ 0.0268 & 0.5777 $\pm$ 0.0285 & 0.7258 $\pm$ 0.0336 \\
ST-Transformer  & 2.6M    & 0.6631 $\pm$ 0.0173 & 0.5672 $\pm$ 0.0259 & 0.7132 $\pm$ 0.0174 \\
\midrule
BIOT            & 3.2M  & 0.6875 $\pm$ 0.0186 & 0.6004 $\pm$ 0.0195 & 0.7536 $\pm$ 0.0144 \\
LaBraM-Base     & 5.8M  & 0.6909 $\pm$ 0.0125 & 0.5999 $\pm$ 0.0155 & 0.7721 $\pm$ 0.0093 \\
CBraMod         & 4.0M  & 0.7256 $\pm$ 0.0132 & 0.6267 $\pm$ 0.0099 & 0.7905 $\pm$ 0.0073 \\
REVE-Base       & 69M   & 0.7660 $\pm$ 0.0355 & 0.7470 $\pm$ 0.0807 & 0.8450 $\pm$ 0.0514 \\
\midrule
BrainCodes & 1.9M & \textbf{0.7720} $\pm$ 0.0143 & \textbf{0.7697} $\pm$ 0.0224 & \textbf{0.8857} $\pm$ 0.0188 \\
\bottomrule
\end{tabular}
\end{center}
\end{table}

\textbf{EESM23}~\citep{bjarke2025ear}. The ear-EEG sleep monitoring dataset contains overnight recordings from 10 participants with electrodes in both ear canals. The four in-ear electrodes were assigned the coordinates of TP9, T7, TP10 and T8, the nearest positions of the standard montage. Following TFM-Tokenizer~\citep{tfmtokenizer}, samples were filtered between 0.1 and 75\,Hz with a 50\,Hz notch, cut into 30-second epochs labeled as wake, N1, N2, N3, REM or artifact, and split so that subjects 1 to 6 were used for training, 7 and 8 for validation and 9 and 10 for testing.

\begin{table}[ht]
\caption{Model performance on low-resource sleep staging (EESM23, 6-class).}
\label{tab:eesm}
\begin{center}
\begin{tabular}{lrccc}
\toprule
\textbf{Models} & \textbf{Size} & \textbf{Balanced Accuracy} & \textbf{Cohen's $\kappa$} & \textbf{Weighted F1} \\
\midrule
BIOT          & 3.2M & 0.3858 $\pm$ 0.0085 & 0.3406 $\pm$ 0.0096 & 0.4888 $\pm$ 0.0124 \\
BIOT-TFM      & 3.2M   & 0.3952 $\pm$ 0.0170 & 0.3603 $\pm$ 0.0252 & 0.5033 $\pm$ 0.0165 \\
LaBraM-Base   & 5.8M & 0.3890 $\pm$ 0.0182 & 0.3322 $\pm$ 0.0232 & 0.4827 $\pm$ 0.0157 \\
LaBraM-TFM    & 5.8M   & 0.4004 $\pm$ 0.0086 & 0.3475 $\pm$ 0.0128 & 0.4864 $\pm$ 0.0118 \\
TFM-Tokenizer & 1.9M & 0.4148 $\pm$ 0.0209 & 0.3883 $\pm$ 0.0233 & 0.5174 $\pm$ 0.0141 \\
\midrule
\textit{Reproduced:} & & & & \\
LaBraM-Base   & 5.8M  & 0.4109 $\pm$ 0.0087 & 0.3589 $\pm$ 0.0084 & 0.4968 $\pm$ 0.0055 \\
CBraMod       & 4.0M  & 0.3993 $\pm$ 0.0163 & 0.3439 $\pm$ 0.0097 & 0.4789 $\pm$ 0.0182 \\
CSBrain       & 8.9M  & 0.3663 $\pm$ 0.0261 & 0.3048 $\pm$ 0.0281 & 0.4642 $\pm$ 0.0191 \\
REVE-Base     & 69M & 0.4176 $\pm$ 0.0224 & 0.3698 $\pm$ 0.0170 & 0.4905 $\pm$ 0.0126 \\
\midrule
BrainCodes & \textbf{1.9M} & \textbf{0.5254} $\pm$ 0.0251 & \textbf{0.4847} $\pm$ 0.0411 & \textbf{0.5908} $\pm$ 0.0268 \\
\bottomrule
\end{tabular}
\end{center}
\end{table}

\textbf{Muse-Motor}~\citep{elkaffas2024musemotor}. The dataset contains recordings from the four dry electrodes of the Muse headband, i.e., TP9, AF7, AF8 and TP10, collected from 19 healthy participants (7 females) over 30 recording sessions and labeled into three motor classes, i.e., left and right motor imagery, and relaxed control. The recordings were band-pass filtered between 0.5 and 75\,Hz with a 60\,Hz notch, resampled to 250\,Hz, robust-z-scored per channel, and segmented into 4-second windows.

\textbf{Muse-Mental}~\citep{bird2018study}. The dataset contains recordings from 4 participants wearing the Muse headband, each recorded while relaxed, neutral and concentrating, and the task was the three-way classification of these mental states. Preprocessing followed Muse-Motor, and segmentation into 4-second windows gave 296 windows evaluated with four-fold cross-validation. Hence, each test fold held about 74 windows, which accounted for the spread across folds in Table~\ref{tab:wearable}.

\begin{table}[ht]
\caption{Detailed results (over 4 cross-validation folds) of BrainCodes on the Muse tasks.}
\label{tab:muse_detailed}
\begin{center}
\begin{tabular}{lccc}
\toprule
\textbf{Dataset} & \textbf{Balanced Accuracy} & \textbf{Cohen's $\kappa$} & \textbf{Weighted F1} \\
\midrule
Muse-Motor (3-class)  & 0.5885 $\pm$ 0.0318 & 0.3650 $\pm$ 0.0516 & 0.5800 $\pm$ 0.0298 \\
Muse-Mental (3-class) & 0.8160 $\pm$ 0.0957 & 0.7097 $\pm$ 0.1287 & 0.8125 $\pm$ 0.0793 \\
\bottomrule
\end{tabular}
\end{center}
\end{table}

\section{Description of Referenced Models}
\label{app:models}

\textbf{EEGNet}~\citep{lawhern_eegnet_2018}: A Compact CNN specifically designed for BCI applications, which uses depthwise and separable convolutions to learn temporal and spatial features from EEG signals.

\textbf{EEGConformer}~\citep{song2022eeg}: A compact model that extracts local temporal and spatial features with convolutions and models global dependencies with self-attention.

\textbf{SPaRCNet}~\citep{jing2023development}: A 1D CNN for EEG decoding that leverages dense residual connections with adaptive depth and channel selection to capture spatiotemporal patterns efficiently.

\textbf{ContraWR}~\citep{yang2023self}: Converts raw EEG into multi-channel time-frequency spectrograms and classifies them with ResNet-style 2D CNN.

\textbf{CNN-Transformer}~\citep{peh2022transformer}: A hybrid model that applies a transformer encoder to features extracted by a convolutional network.

\textbf{ST-Transformer}~\citep{song2021transformer}: A transformer-based neural network that applies attention separately over the spatial and the temporal dimension of EEG.

\textbf{BIOT}~\citep{biot}: A linear Transformer for universal biosignal representation learning that uses hybrid supervised-unsupervised pretraining and projects inputs with more than 18 EEG channels into an 18-channel format via a $1\times1$ convolution.

\textbf{LaBraM}~\citep{labram}: An EEG foundation model that employs a full-attention Transformer to capture generalizable neural representations through masked neural token prediction.

\textbf{CBraMod}~\citep{cbramod}: An EEG foundation model that leverages a criss-cross Transformer backbone to capture the heterogeneous spatiotemporal structure of EEG signals.

\textbf{CSBrain}~\citep{csbrain}: An attention-based foundation model for EEG decoding with novel cross-scale spatiotemporal tokenization and structured sparse attention.

\textbf{REVE}~\citep{reve}: A masked-autoencoder EEG foundation model pretrained on more than 60,000 hours from 25,000 subjects, which encodes electrode coordinates and time jointly through a four-dimensional positional encoding.

\textbf{EEGPT}~\citep{eegpt}: A pretrained EEG transformer trained with masked reconstruction together with an alignment of spatio-temporal representations.

\textbf{NeuroGPT}~\citep{neurogpt}: A foundation model that pairs an EEG encoder with a causal GPT-style transformer, pretrained on TUEG to predict masked segments from the preceding ones.

\textbf{BrainOmni}~\citep{brainomni}: A foundation model for EEG and MEG that maps heterogeneous sensor layouts into a shared discrete token space before transformer pretraining.

\textbf{NeuroRVQ}~\citep{neurorvq}: A generative EEG foundation model built on a multi-scale residual vector-quantized tokenizer, which also introduced the benchmark suite of Appendix~\ref{app:neurorvq}.

\textbf{TFM-Tokenizer}~\citep{tfmtokenizer}: A single-channel EEG tokenizer that learns a vocabulary of time-frequency motifs, used either with its own transformer or as a front end to existing models such as BIOT and LaBraM.

\textbf{DeepSleepNet}~\citep{supratak2017deepsleepnet}: A sleep staging model for raw single-channel EEG that learns epoch-level representations with convolutional networks and models the transitions between consecutive epochs with bidirectional long short-term memory networks.

\textbf{U-Sleep}~\citep{perslev2021u}: A fully convolutional encoder-decoder for sleep staging, trained and evaluated on polysomnography from 15,660 participants of 16 clinical studies. It accepts any combination of the standard EEG and EOG channels, and although trained on 30-second epochs it can output sleep stages at a finer temporal resolution.

\section{Implementation Details \& Resources}

\textbf{Pretraining.} BrainCodes was pretrained on the codes of the frozen BioCodec tokenizer, i.e., six residual quantization stages of 256 codewords at 20.8 frames per second, extracted from the corpus described in Section~3 after band-pass filtering between 0.5 and 75\,Hz with a 60\,Hz notch, interpolation of flat channels, resampling to 250\,Hz and robust z-scoring of each channel. Training crops were drawn with equal probability among lengths of 2, 5, 10, 15, 20 and 30\,s, and each crop retained a random subset of 16 to 64 channels, so that the model encountered both short windows and sparse montages during pretraining. The backbone used a model width of 128 with 8 attention heads, three blocks of two time-attention layers and one channel-attention layer, temporal pooling by factors of 2 and 2, encoder dropout of 0.1 and U-Net skip-connection dropout of 0.2. Optimization used AdamW with $\beta$ = (0.9, 0.95), a peak learning rate of 5\texttimes{}10\textsuperscript{\textminus 4}, weight decay of 0.05, 2,000 warmup steps followed by cosine decay and gradient clipping at 1.0, for 20,000 steps at a batch size of 256 samples, realized through gradient accumulation in bfloat16. Each batch was assigned to one of the two objectives, i.e., 60\% to masked code prediction and 40\% to oscillation injection, with unit weight on every loss term. The block mask selected between 10\% and 25\% of channels, gave each a contiguous span whose length was drawn log-uniformly between 20\% and 80\% of the batch length, and extended every span to the 12.5\% of electrodes nearest to that channel, so that a masked code could not be copied from an adjacent contact. Injected bursts had amplitudes between 0.1 and 0.5, a bandwidth of 2\,Hz, durations between 10\% and 50\% of the batch length, and were applied to regions of 20\% to 40\% of the channels; the location head was trained with binary cross-entropy and the frequency and duration heads with squared error on unit-normalized targets. Electrode coordinates were further perturbed with Gaussian noise of standard deviation 0.05, capped at half the mean nearest-neighbor spacing of the montage. Every downstream experiment used the checkpoint at step 20,000. Pretraining required about 190 GPU-hours, realized on 4 NVIDIA RTX 6000 GPUs.

\textbf{Downstream Adaptation.} Unless stated otherwise, BrainCodes was fully fine-tuned with AdamW at a weight decay of 0.01, label smoothing of 0.1 and class weights set to the square root of the inverse class frequency, normalized to unit mean, which damps the weighting of rare classes relative to inverse-frequency weights. Inputs were z-scored per channel and window, the learning rate rose linearly during the warmup epochs and then followed a cosine decay, and the checkpoint with the highest validation balanced accuracy was evaluated on the test data. The classification head read either the time-pooled bottleneck feature $b_c$ through a projector shared across channels or the flattened bottleneck $\bar{b}_c$ through a single linear layer, and the per-dataset settings are listed in Table~\ref{tab:hparams}.

\begin{table}[t]
\caption{Fine-tuning settings of BrainCodes for the results reported in the main benchmarks.}
\label{tab:hparams}
\begin{center}
\begin{tabular}{lcccr}
\toprule
\textbf{Dataset} & \textbf{Head} & \textbf{Peak LR} & \textbf{Epochs (warmup)} & \textbf{Batch} \\
\midrule
TUAB & $\bar{b}_c$ & 5\texttimes{}10\textsuperscript{\textminus 4} & 10 (2) & 128 \\
TUEV & $\bar{b}_c$ & 5\texttimes{}10\textsuperscript{\textminus 4} & 10 (2) & 32 \\
Siena & $b_c$ & 5\texttimes{}10\textsuperscript{\textminus 4} & 20 (2) & 32 \\
ISRUC & $b_c$ & 5\texttimes{}10\textsuperscript{\textminus 4} & 20 (2) & 64 \\
PhysioNet-MI & $b_c$ & 5\texttimes{}10\textsuperscript{\textminus 4} & 20 (2) & 64 \\
BCI IV-2a & $\bar{b}_c$ & 2\texttimes{}10\textsuperscript{\textminus 4} & 20 (2) & 32 \\
EEGMAT & $b_c$ & 5\texttimes{}10\textsuperscript{\textminus 4} & 20 (2) & 32 \\
\midrule
High-Gamma & $b_c$ & 5\texttimes{}10\textsuperscript{\textminus 4} & 20 (2) & 64 \\
OpenBMI-ERP & $b_c$ & 5\texttimes{}10\textsuperscript{\textminus 4} & 20 (4) & 64 \\
Pavlov-Memory & $b_c$ & 5\texttimes{}10\textsuperscript{\textminus 4} & 20 (2) & 64 \\
PhysioNet-Eyes & $b_c$ & 5\texttimes{}10\textsuperscript{\textminus 4} & 20 (2) & 16 \\
\midrule
EESM23 & $b_c$ & 1\texttimes{}10\textsuperscript{\textminus 4} & 20 (2) & 64 \\
Muse-Motor & $b_c$ & 5\texttimes{}10\textsuperscript{\textminus 4} & 20 (4) & 16 \\
Muse-Mental & $b_c$ & 5\texttimes{}10\textsuperscript{\textminus 4} & 40 (4) & 16 \\
\bottomrule
\end{tabular}
\end{center}
\end{table}

\textbf{Baseline Reproduction.} The baselines of the consumer-hardware benchmark were run on the per-fold splits of BrainCodes. On the Muse tasks, every model was fine-tuned with AdamW at a learning rate of 5\texttimes{}10\textsuperscript{\textminus 4}, weight decay of 0.01, label smoothing of 0.1 and square-root class weights, for 20 epochs on Muse-Motor and 40 on Muse-Mental. On EESM23 the reproduced baselines of Table~\ref{tab:eesm} were fine-tuned on the data build of BrainCodes with its settings. BIOT was omitted from the Muse tasks, since its bipolar derivations could not be derived from the four Muse electrodes.

\textbf{Grounding-probe implementation.} Both probes in Section~\ref{sec:suite} used the test split of PhysioNet-MI under the standard BrainCodes preprocessing, with 128 windows for the spatial probe and 512 for the spectral probe. For the spatial probe, we perturbed one electrode at a time by adding Gaussian white noise at an SNR of 10\,dB and measured its influence on channel $i$ as $\lVert f_i(\tilde{x}) - f_i(x) \rVert / \lVert f_i(x) \rVert$, where $f_i$ denotes the features of channel $i$ and the norms run over windows and feature dimensions. Mixing was defined as the share of the total influence carried by pairs of distinct electrodes, locality as the Spearman correlation between that influence and inter-electrode distance, and the decay length $\lambda$ as $-1/b$ for the slope $b$ of a least-squares fit of log influence against Euclidean distance, with coordinates centered on the montage and scaled to a maximum electrode norm of 1. For the spectral probe, band-limited noise was injected into a spatially contiguous 20--40\% of the channels, i.e., a seed electrode and its nearest neighbors, as delta (0.5 to 4\,Hz) and spindle (11 to 16\,Hz) bursts lasting 0.5 to 2.0 and 0.5 to 1.5\,s under Hann and Gaussian envelopes, and as continuous mu (8 to 13\,Hz, central electrodes only) and beta (13 to 30\,Hz) activity, while white and pink noise were matched in power; the SNR of 35 to 15\,dB was defined as the ratio of signal to marker power per injected channel over the injection window, before any normalization. Standardized logistic regressions were fitted to the change in channel-averaged features between the injected and the clean window, reduced to 64 principal components, under leave-one-participant-out validation.

\textbf{Computational Cost.} BrainCodes holds 1,877,312 trainable parameters, and the BioCodec tokenizer it reads from contributes a further 3,152,770 that remained frozen throughout, so the released checkpoint stores the encoder alone in 7.19\,MiB. With its classification head the fine-tuned model held between 1.9 and 2.5 million trainable parameters across the datasets of Table~\ref{tab:hparams}, from 1,917,510 on EESM23 to 2,520,744 on TUAB. Tables with results only mention the size of the encoders. Because the tokenizer was frozen and shared across every downstream task, each corpus was tokenized once and its code indices were cached. Hence the fine-tuning runs of Appendix~\ref{app:budget} and the arms of Table~\ref{tab:withheld} shared a single tokenization pass per corpus.

\textbf{Software and Hardware.} Pretraining, fine-tuning, and analyses ran on a dedicated cluster of 4 RTX 6000 GPUs, in Python 3.12 with PyTorch 2.11~\citep{paszke2019pytorch}, scikit-learn 1.9 and MNE-Python~\citep{gramfort2013meg}. Electrode coordinates were resolved against its \texttt{standard\_1005} montage. Further details on hyperparameters are provided with the associated GitHub repository, which will be made publicly available, including code for model build, training, and model weights.

\end{document}